\documentclass{article}
\usepackage[preprint]{colm2026_conference}

\usepackage{microtype}
\usepackage{hyperref}
\usepackage{url}
\usepackage{booktabs}
\usepackage{amsmath}
\usepackage{amssymb}
\usepackage{graphicx}
\usepackage{multirow}
\usepackage{xcolor}
\usepackage{subcaption}
\usepackage{enumitem}
\usepackage{amsthm}

\usepackage{lineno}
\usepackage{framed}

\definecolor{darkblue}{rgb}{0, 0, 0.5}
\definecolor{promptframe}{RGB}{30, 70, 130}
\definecolor{promptbg}{RGB}{240, 245, 252}
\definecolor{rubricbg}{RGB}{248, 248, 240}
\hypersetup{colorlinks=true, citecolor=darkblue, linkcolor=darkblue, urlcolor=darkblue}

\newenvironment{promptbox}[1][Prompt]{%
  \par\medskip
  \noindent\colorbox{promptframe}{\color{white}\bfseries\small\strut\,#1\,}\par
  \vspace{-0.55\baselineskip}
  \colorlet{shadecolor}{promptbg}%
  \begin{shaded}\noindent\ignorespaces
}{%
  \end{shaded}\medskip
}

\title{VLM Judges Can Rank but Cannot Score: Task-Dependent Uncertainty in Multimodal Evaluation}

\author{Divake Kumar$^{1}$, Sina Tayebati$^{1}$, Devashri Naik$^{1}$, Ranganath Krishnan$^{2}$, Amit Ranjan Trivedi$^{1}$ \\[4pt]
$^{1}$University of Illinois at Chicago, $^{2}$AI Labs at Capital One
}

\begin{document}

\ifcolmsubmission
\linenumbers
\fi

\maketitle

\begin{abstract}
Vision-language models (VLMs) are increasingly used as automated judges for multimodal systems, yet their scores provide no indication of reliability. We study this problem through conformal prediction, a distribution-free framework that converts a judge's point score into a calibrated prediction interval using only score-token log-probabilities, with no retraining. 
We present the first systematic analysis of conformal prediction for VLM-as-a-Judge across 3 judges and 14 visual task categories. Our results show that evaluation uncertainty is strongly task-dependent: intervals cover $\sim 40\%$ of the score range for aesthetics and natural images but expand to $\sim 70\%$ for chart and mathematical reasoning, yielding a quantitative reliability map for multimodal evaluation. 
We further identify a failure mode not captured by standard evaluation metrics, \emph{ranking-scoring decoupling}, where judges achieve high ranking correlation while producing wide, uninformative intervals, correctly ordering responses but failing to assign reliable absolute scores. 
Finally, we show that interval width is driven primarily by task difficulty and annotation quality, i.e., the same judge and method yield $4.5\times$ narrower intervals on a clean, multi-annotator captioning benchmark. \\[4pt]
\textbf{Code:} \href{https://github.com/divake/VLM-Judge-Uncertainty}{\texttt{github.com/divake/VLM-Judge-Uncertainty}}
\end{abstract}

\section{Introduction}
\label{sec:intro}

Vision-Language Models (VLMs) are increasingly used as automated judges to evaluate multimodal AI systems~\citep{chen2024mllmasajudge,li2024vlrewardbench}. This \emph{VLM-as-a-Judge} paradigm extends LLM-based evaluation to tasks requiring visual understanding, including chart comprehension, visual question answering, image captioning, and AI-generated image quality. As multimodal agents are deployed in high-stakes settings such as medical imaging, web navigation, and document analysis, the reliability of these evaluations becomes critical, paralleling broader concerns about uncertainty quantification and robustness in safety-critical autonomous systems~\citep{trivedi2025sensing,darabi2024navigating,darabi2025intact,poggi2026depth,naik2026belief}.

However, VLM judges produce only point scores, offering no indication of reliability. A score of four out of five (4/5) may reflect confident correctness or substantial uncertainty, yet users cannot distinguish between the two. This limitation is exacerbated in multimodal settings, where errors can arise from both visual perception and language reasoning. For example, evaluating a chart-based answer requires visual parsing, data extraction, and reasoning, each introducing compounding uncertainty. Empirically, this reliability gap is significant. On the MLLM-as-a-Judge benchmark~\citep{chen2024mllmasajudge}, state-of-the-art VLM judges achieve only 32--34\% exact agreement with human ratings on a 5-point scale, and 24--30\% of predictions deviate by 2 or more points. Correlation metrics remain modest ($\rho = 0.30$--$0.46$), providing limited assurance about per-instance reliability. Existing evaluation metrics thus fail to indicate when a judge's score can be trusted.

We address this problem using conformal prediction~\citep{vovk2005algorithmic}, a post-hoc, distribution-free framework that converts point scores into calibrated prediction intervals with provable coverage guarantees. Given a target coverage (e.g., 90\%), these intervals contain the true human score with high probability, without requiring model retraining or assumptions on error distributions. Conformal prediction operates directly on score-token log-probabilities, making it readily applicable to existing VLM judge pipelines. While prior work~\citep{sheng2025analyzing} demonstrates the effectiveness of conformal prediction for text-only LLM judges, extending it to multimodal evaluation introduces new challenges: (i) cross-modal uncertainty from visual and textual reasoning, (ii) discrete single-annotator ground truth, (iii) substantial task heterogeneity across visual domains, and (iv) varying levels of log-probability access across models. Standard metrics such as correlation and accuracy fail to capture evaluation reliability. 

We present a systematic study of conformal prediction for VLM-based evaluation. Our contribution is not the application of conformal prediction itself, but the identification of task-dependent uncertainty structure and a fundamental decoupling between ranking quality and scoring reliability in multimodal evaluation. We show that high correlation does not imply reliable absolute scores, and that uncertainty is primarily driven by task ambiguity and annotation quality rather than model scale. Our key findings are:

\begin{itemize}[leftmargin=*,itemsep=0pt,topsep=0pt]

\item \textbf{Task-dependent uncertainty:} Prediction interval width varies by up to 70\% across tasks. Vision-heavy tasks such as charts and mathematical reasoning yield substantially wider intervals than natural image or aesthetics tasks, providing a quantitative reliability map.

\item \textbf{Ranking-scoring decoupling:} We identify a failure mode not captured by standard evaluation metrics where judges achieve strong ranking correlation while producing wide, uninformative intervals. That is, they correctly order responses but fail to assign reliable absolute scores.

\item \textbf{Role of data quality:} Interval width is primarily driven by task difficulty and annotation quality rather than the conformal method. The same judge and method produce $4.5\times$ narrower intervals on a clean, multi-annotator captioning benchmark.

\item \textbf{Task-conditional calibration:} We show that conditioning on task difficulty (Mondrian CP) yields narrower intervals for easy tasks while improving coverage for hard tasks. \vspace{-3pt}

\end{itemize}

\section{Related work}
\label{sec:related}

\textbf{LLM-as-a-Judge.}
Using LLMs as automated evaluators has become a standard approach for assessing natural language generation~\citep{zheng2023judging,liu2023geval}. G-Eval~\citep{liu2023geval} uses chain-of-thought prompting with GPT-4 to score text quality and improves correlation with human judgments over metrics such as ROUGE and BERTScore. MT-Bench and Chatbot Arena~\citep{zheng2023judging} establish pairwise evaluation and reveal systematic biases including position and verbosity bias. SocREval~\citep{he2024soceval} focuses on reasoning. 

\textbf{VLM-as-a-Judge.}
This paradigm has been extended to multimodal evaluation. \citet{chen2024mllmasajudge} introduce the MLLM-as-a-Judge benchmark with 14 visual task categories and report limited alignment with human judgments. LLaVA-Critic~\citep{xiong2024llavacritic} trains a dedicated multimodal judge, while VL-RewardBench~\citep{li2024vlrewardbench} shows that a large fraction of errors are perceptual rather than reasoning-driven. 

\textbf{Uncertainty in LLM-based evaluation.}
Recent work studies confidence and uncertainty in LLM judgments. \citet{wagner2024llmconfidence} derive confidence scores from token probabilities and show weak correlation with accuracy. \citet{xie2024boosted} use token probabilities for calibration and propose boosted conformal methods. Other approaches rely on self-reported confidence~\citep{xiong2024selfconsistency} or consistency across multiple generations~\citep{tian2023consistency}, both with known limitations. \citet{sheng2025analyzing} apply conformal prediction to text-only LLM judges and show that calibrated intervals improve reliability. Beyond pointwise calibration, recent work studies conformalized abstention policies that selectively defer uncertain predictions in LLM and VLM deployments~\citep{tayebati2025conformalabstention,tayebati2025cap}, and trajectory-level risk aggregation for agentic reasoning~\citep{tayebati2026tracer}; our work is complementary in that it directly characterizes the score-token uncertainty of the judge itself rather than acting on it.

\textbf{Conformal prediction.}
Conformal prediction provides distribution-free uncertainty estimates with coverage guarantees~\citep{vovk2005algorithmic}. Prior work applies it mainly to classification tasks such as multiple-choice QA~\citep{kumar2023conformal,su2024api}, factual consistency~\citep{quach2024conformal,mohri2024language}, and calibrated risk control~\citep{angelopoulos2021conformal}. Recent extensions move beyond fixed nonconformity scores: \citet{kumar2025learnablecp} learn context-aware nonconformity functions for robotic planning and perception, \citet{stutts2024conformal} couple conformal inference with evidential learning for epistemic--aleatoric separability, and follow-up work decomposes and routes the two uncertainty components for inference-time adaptation and intervention~\citep{kumar2025calibrated,kumar2026triage}. Conformal prediction has likewise been used to enable principled abstention in safety-critical multi-sensor autonomy~\citep{kumar2025lidarcp}.

\section{Uncertainty Quantification for VLM-based Judge}
\label{sec:uncertainty}

\subsection{VLM-as-a-Judge framework}

In the VLM-as-a-Judge setting, a vision-language model $M$ evaluates the quality of a response given an image $I$ and question $q$. For a candidate answer $a$, the model produces a score $y_0$ on a discrete Likert scale $\{1, \ldots, K\}$ along with token logits:
\begin{equation}
M(I, q, a) = (z, y_0).
\end{equation}
We extract the log-probabilities of each rating token at the score position, forming a $K$-dimensional feature vector:
\begin{equation}
\mathbf{x} = [\log P(\text{``1''}), \log P(\text{``2''}), \ldots, \log P(\text{``$K$''})].
\label{eq:features}
\end{equation}
This feature vector captures the model's distribution over possible scores. A concentrated distribution, where one score has high probability, indicates high confidence, while a diffuse distribution indicates uncertainty. We also considered additional features derived from entropy, margin of the score distribution, and actor-model entropy. These did not improve performance over $\mathbf{x}$ alone, so we use this representation throughout. 

\subsection{Conformal prediction}

Conformal prediction~\citep{vovk2005algorithmic} is a distribution-free framework for constructing prediction intervals with coverage guarantees. Given a calibration set $\{(\mathbf{x}_i, y_i)\}_{i=1}^n$ and a nonconformity score function $s(\mathbf{x}, y)$, we compute the quantile $\hat{q}$ at miscoverage level $\alpha$. For a test point $\mathbf{x}$, the prediction interval is
\begin{equation}
\mathcal{C}(\mathbf{x}) = [\hat{y} - \hat{q},\ \hat{y} + \hat{q}],
\end{equation}
where $\hat{y} = f(\mathbf{x})$ is a point predictor trained on the calibration set. Under the exchangeability assumption, this interval satisfies
\begin{equation}
\mathbb{P}(y \in \mathcal{C}(\mathbf{x})) \geq 1 - \alpha.
\end{equation}
Conformal prediction is well-suited to this setting because it is post-hoc and does not require retraining or fine-tuning of the judge model, and it is distribution-free, avoiding assumptions on error distributions that may vary across multimodal tasks.

\subsection{Extensions and diagnostics}

\paragraph{Boundary adjustment for discrete scales.}
Ground truth scores are discrete integers, while conformal prediction produces continuous intervals. Following \citet{sheng2025analyzing}, we apply boundary adjustment:
\begin{equation}
s'(\mathbf{x}, y) =
\begin{cases}
s(\mathbf{x}, \lceil y \rceil) & \text{if } y \leq \lfloor \hat{y} \rfloor, \\
s(\mathbf{x}, y) & \text{otherwise,} \\
s(\mathbf{x}, \lfloor y \rfloor) & \text{if } y \geq \lceil \hat{y} \rceil,
\end{cases}
\end{equation}
which maps the interval $[l, u]$ to $[\lceil l \rceil, \lfloor u \rfloor]$, aligning endpoints with valid rating labels while preserving or increasing coverage.

\paragraph{Task-conditional conformal prediction.}
Standard conformal prediction uses a single global quantile. To account for task-dependent variability, Mondrian conformal prediction partitions data into groups and computes group-specific quantiles, yielding group-conditional guarantees:
\begin{equation}
\mathbb{P}(y \in \mathcal{C}(\mathbf{x}) \mid G = g) \geq 1 - \alpha.
\end{equation}
This produces tighter intervals for easier tasks and more conservative intervals for harder. We partition the 14 MLLM-Judge datasets into three groups based on per-dataset R2CCP width from an initial standard CP run: \emph{easy} (width ${<}\,2.5$: AesBench, MM-Vet, WIT, COCO), \emph{medium} ($2.5$--$3.2$: Mind2Web, Conceptual Captions, TextVQA, LLaVA-Bench, VisitBench, ChartQA), and \emph{hard} (${>}\,3.2$: ScienceQA, MathVista, DiffusionDB, InfographicsVQA). Group boundaries are determined once on a held-out pilot split and fixed for all subsequent experiments.

\paragraph{Ranking-Scoring Gap.}
Standard metrics capture ranking quality but not the reliability of absolute scores. We define the Ranking-Scoring Gap:
\begin{equation}
\text{RSG}(d) = |\rho_d| - \left(1 - \frac{w_d}{K - 1}\right),
\end{equation}
where $\rho_d$ is the Pearson correlation and $w_d$ is the average interval width. The second term measures interval informativeness. High RSG indicates strong ranking but unreliable absolute scores. Following \citet{sheng2025analyzing}, we also evaluate the midpoint $\hat{y}_{\text{mid}} = (l + u)/2$ as a calibrated estimate and assess its effectiveness in multimodal settings.

\section{Experimental setup}
\label{sec:setup}

\textbf{Datasets and Judges.}
We use two multimodal evaluation benchmarks with human-annotated ground truth scores.

\textbf{(i) MLLM-as-a-Judge~\citep{chen2024mllmasajudge}.}
This benchmark contains 5,717 instances spanning 14 visual task categories. Each instance includes an image, a question, a model-generated answer, and a human-annotated score on a 1--5 Likert scale from a single annotator. It covers diverse tasks including visual reasoning, chart understanding, web-based queries, and aesthetic evaluation. The score distribution is skewed toward higher values, which affects calibration and conditional coverage.

\textbf{(ii) Polaris~\citep{wada2024polos}.}
This captioning benchmark contains 8,726 image-caption pairs with scores aggregated from multiple annotators and mapped to a 1--5 scale. Compared to MLLM-as-a-Judge, Polaris represents a single, well-defined task with lower annotation noise, allowing isolation of the effects of task difficulty and label quality on interval width.

We evaluate three VLM judges spanning different architectures, scales, and access settings: \textbf{(i) LLaVA-Critic-7B~\citep{xiong2024llavacritic}}, an open-source 7B model specialized for multimodal evaluation; \textbf{(ii) Phi-4-reasoning-vision-15B~\citep{abdin2025phi4reasoning}}, a 15B reasoning model with extended chain-of-thought generation; and \textbf{(iii) Gemini 2.5 Flash~\citep{google2025gemini25}}, a closed-source API model with strong ranking performance but highly confident predictions. All models use a consistent evaluation prompt across datasets (Appendix~\ref{app:prompts}), enabling controlled comparison across architectures.

\textbf{Evaluation protocol.}
Following \citet{sheng2025analyzing}, we randomly split each dataset 50/50 into calibration and test sets across 10 random seeds and report mean $\pm$ standard deviation. We use significance level $\alpha = 0.10$, targeting 90\% coverage. This yields approximately 2,859 calibration and 2,858 test samples per split for MLLM, and 4,363/4,363 for Polaris. For each instance, we extract score-token log-probabilities to form a 5-dimensional feature vector. Additional implementation details are provided in Appendix~\ref{app:feature_extraction}.

We report the following metrics. Coverage measures the fraction of test samples where the ground truth lies within the prediction interval, with a target of $\geq 0.90$. Width is the average interval size on the 1--5 scale (range 0 to 4), where lower is better given sufficient coverage. Point prediction quality is evaluated using Pearson $\rho$, Spearman $\rho_s$, and Kendall $\tau$ correlations, along with exact-match accuracy, relaxed $\pm 1$ accuracy, and mean absolute error. We also report judge bias as the mean signed error (judge score minus human ground truth), stratified by ground truth level. Both raw (continuous) and boundary-adjusted (discrete) intervals are reported for all methods. For per-dataset analysis, we run R2CCP independently on each of the 14 datasets to evaluate conditional coverage and task-dependent interval widths.

\section{Results and Discussions}

Before examining judge uncertainty, we establish baseline point prediction performance for all three judges on MLLM-as-a-Judge (Table~\ref{tab:cross_judge}). All three judges achieve comparable exact accuracy (32--34\%) but differ substantially in other metrics. Gemini achieves the highest correlation (Pearson $= 0.459$), Phi-4 the lowest MAE (1.005) and lowest bias ($+0.08$), and LLaVA-Critic produces the most calibrated log-probability distributions (25.6\% overconfident). These differences directly affect the quality of VLM-judge uncertainty prediction, shaping both interval width and coverage.

\begin{table}[t]
\centering
\caption{Cross-judge comparison on MLLM-as-a-Judge (10 seeds). Metrics include point prediction, score calibration, and conformal prediction results.}
\label{tab:cross_judge}
\small
\begin{tabular}{llccc}
\toprule
\textbf{Group} & \textbf{Metric} & \textbf{LLaVA-Critic 7B} & \textbf{Phi-4 15B} & \textbf{Gemini 2.5 Flash} \\
\midrule

Point pred. & Pearson $\rho$ & .402 & .303 & \textbf{.459} \\
 & Spearman $\rho_s$ & .356 & .293 & \textbf{.446} \\
 & Kendall $\tau$ & .300 & .251 & \textbf{.377} \\
 & Accuracy & 32.2\% & \textbf{34.2\%} & 32.1\% \\
 & $\pm 1$ Acc. & 75.1\% & \textbf{76.1\%} & 70.3\% \\
 & MAE & 1.031 & \textbf{1.005} & 1.122 \\
 & Bias & $+0.38$ & $+0.08$ & $\mathbf{-0.05}$ \\
\midrule
Score cal. & Overconf. ($>$0.99) & \textbf{25.6\%} & 43.0\% & 83.0\% \\
 & Overconf. ($>$0.999) & \textbf{6.6\%} & 16.9\% & 77.4\% \\
\midrule
R2CCP & Coverage (raw) & \textbf{.900} & .891 & .898 \\
 & Width (raw) & 3.05 & 3.13 & \textbf{2.85} \\
 & Coverage (adj.) & .981 & .981 & .980 \\
 & Width (adj.) & 3.60 & 3.70 & \textbf{3.41} \\
\midrule
CHR & Coverage (raw) & .880 & .896 & .880 \\
 & Width (raw) & 2.97 & 3.16 & \textbf{2.80} \\

\bottomrule\vspace{-15pt}
\end{tabular}
\end{table}

\subsection{Conformal methods comparison}
\label{sec:methods_comparison}

Table~\ref{tab:all_methods} reports coverage and interval width for all 8 conformal prediction methods on MLLM-as-a-Judge using LLaVA-Critic-7B. The methods fall into three performance tiers.

R2CCP achieves coverage closest to the 90\% target (90.0\%) with width 3.05, providing the best balance between calibration and interval size. CHR produces slightly narrower intervals (2.97) but under-covers (88.0\%), while Boosted LCP yields stable intervals but also under-covers. After boundary adjustment, all three exceed 96\% coverage. In contrast, CQR, Asymmetric CQR, and OrdinalAPS collapse to near-full-range intervals (${\sim}4.0$), achieving near-100\% coverage but providing little discrimination. After boundary adjustment, CQR variants produce constant intervals $[1,5]$ for all samples.

This difference arises from how methods use the input features. Approaches based on conditional quantile estimation degrade when features are weak predictors, leading to uniformly wide intervals. In contrast, R2CCP models the full conditional distribution $P(y \mid \mathbf{x})$, which is more robust in this setting. Boundary adjustment is disproportionately important in our setting: it increases coverage by 8+ percentage points on our 5-class integer GT, compared to 2--5 points reported by \citet{sheng2025analyzing} on 13-class averaged GT, because continuous interval endpoints frequently fall between our sparse integer labels. Based on these results, we use \textbf{R2CCP} as the default method due to its accurate coverage and reasonable interval width. CHR is preferred when narrower intervals are desired and slight under-coverage is acceptable.

\begin{table}[t]
\centering
\caption{Conformal prediction methods on MLLM-as-a-Judge (LLaVA-Critic-7B, $\alpha=0.10$, 10 seeds). Methods grouped by performance tier. Bold indicates best within well-calibrated.}
\label{tab:all_methods}
\small
\begin{tabular}{lcccc}
\toprule
\textbf{Method} & \textbf{Cov. (raw)} & \textbf{Width (raw)} & \textbf{Cov. (adj.)} & \textbf{Width (adj.)} \\
\midrule
\multicolumn{5}{l}{\emph{Tier 1: Well-calibrated (coverage $\approx$ 90\%, informative width)}} \\
\textbf{R2CCP} & \textbf{.900$\pm$.016} & 3.05$\pm$.10 & .981$\pm$.005 & 3.60$\pm$.09 \\
CHR & .880$\pm$.010 & \textbf{2.97$\pm$.07} & .965$\pm$.006 & \textbf{3.43$\pm$.10} \\
Boosted LCP & .863$\pm$.012 & 3.02$\pm$.03 & .977$\pm$.004 & 3.51$\pm$.04 \\
\midrule
\multicolumn{5}{l}{\emph{Tier 2: Moderate (slight over/under-coverage, moderate width)}} \\
Naive Split CP & .895$\pm$.009 & 3.23$\pm$.06 & .990$\pm$.003 & 3.78$\pm$.04 \\
LVD & .894$\pm$.009 & 3.21$\pm$.11 & .988$\pm$.005 & 3.71$\pm$.11 \\
Boosted CQR & .878$\pm$.009 & 3.54$\pm$.09 & .997$\pm$.002 & 3.93$\pm$.04 \\
\midrule
\multicolumn{5}{l}{\emph{Tier 3: Degenerate (massive over-coverage, uninformative)}} \\
CQR & .996$\pm$.005 & 3.95$\pm$.06 & 1.00$\pm$.000 & 4.00$\pm$.00 \\
Asym. CQR & .996$\pm$.005 & 3.95$\pm$.06 & 1.00$\pm$.000 & 4.00$\pm$.00 \\
OrdinalAPS & .999$\pm$.000 & 3.99$\pm$.00 & .999$\pm$.000 & 3.99$\pm$.00 \\
\bottomrule\vspace{-15pt}
\end{tabular}
\end{table}

\subsection{Cross-judge comparison}
\label{sec:cross_judge}

Table~\ref{tab:cross_judge} compares the three VLM judges on both point prediction and conformal prediction metrics, revealing several non-obvious patterns.

Despite producing highly overconfident score distributions (83\% of softmax maxima $> 0.99$), Gemini yields the narrowest conformal intervals (2.85 raw, 3.41 adjusted), outperforming both LLaVA-Critic and Phi-4. This apparent paradox highlights a key property of conformal prediction: interval quality depends on the discriminative signal in the feature representation rather than the calibration of the softmax distribution. Even when probabilities are sharply peaked, relative differences between score-token log-probabilities remain informative, enabling R2CCP to construct tight and valid intervals. Overconfidence in raw probabilities therefore does not imply poor uncertainty estimation.

The three judges also exhibit a clear tradeoff between accuracy and correlation. Phi-4 achieves the highest exact accuracy (34.2\%) and lowest MAE (1.005) but the lowest Pearson correlation (0.303), while Gemini achieves the highest correlation (0.459) but lower $\pm 1$ accuracy (70.3\%) and higher MAE (1.122). This divergence reflects two distinct aspects of judge quality: correlation captures ranking ability, while accuracy reflects absolute scoring precision. Gemini is a stronger ranker, whereas Phi-4 is more precise in absolute scoring.

This finding has a concrete mechanism: R2CCP learns a neural network mapping from the 5-dimensional logprob vector to nonconformity scores. Even when Gemini's softmax distribution is sharply peaked, the \emph{relative} magnitudes across the five score tokens still carry discriminative information---for instance, whether the second-highest logprob is for an adjacent score or a distant one. R2CCP exploits these relative patterns regardless of absolute calibration. Overall, conformal interval quality is governed by the signal-to-noise ratio of the representation rather than probability calibration.

\begin{table}[t]
\centering
\caption{Per-dataset R2CCP analysis for LLaVA-Critic-7B on MLLM-as-a-Judge (10 seeds). Datasets sorted by raw width. Width varies 70\% across task types---from 2.08 (AesBench) to 3.50 (InfographicsVQA).}
\label{tab:per_dataset}
\small
\begin{tabular}{lrccccc}
\toprule
\textbf{Dataset} & \textbf{N} & \textbf{Width} & \textbf{Width} & \textbf{Pearson} & $\pm$\textbf{1} & \textbf{Category} \\
& & \textbf{(raw)} & \textbf{(adj.)} & & \textbf{Acc.} & \\
\midrule
AesBench & 392 & \textbf{2.08} & \textbf{2.66} & .402 & 88.3\% & Aesth. \\
MM-Vet & 258 & 2.18 & 2.76 & .260 & 80.2\% & Gen. VQA \\
WIT & 399 & 2.38 & 3.15 & .164 & 78.7\% & Know. \\
COCO & 397 & 2.43 & 3.07 & .362 & 84.9\% & Gen. VQA \\
Mind2Web & 398 & 2.69 & 3.31 & .268 & 74.4\% & Know. \\
Conc. Cap. & 398 & 2.70 & 3.38 & .362 & 81.7\% & Know. \\
TextVQA & 399 & 2.81 & 3.28 & .389 & 79.2\% & Vision \\
LLaVA-B. & 396 & 2.92 & 3.61 & .244 & 81.1\% & Gen. VQA \\
VisitBench & 397 & 2.96 & 3.48 & .352 & 75.1\% & Gen. VQA \\
ChartQA & 400 & 3.08 & 3.57 & \textbf{.507} & 73.2\% & Vision \\
ScienceQA & 396 & 3.27 & 3.64 & .258 & 69.7\% & Vision \\
MathVista & 790 & 3.37 & 3.71 & .376 & 65.1\% & Vision \\
DiffusionDB & 299 & 3.41 & 3.89 & .089 & 57.2\% & Aesth. \\
Infograph. & 398 & \textbf{3.50} & \textbf{3.84} & .411 & 69.3\% & Vision \\
\bottomrule\vspace{-10pt}
\end{tabular}
\end{table}

\subsection{Error structure and bias}
\label{sec:error_bias}

\begin{figure}[t]
\centering
\includegraphics[width=0.8\linewidth]{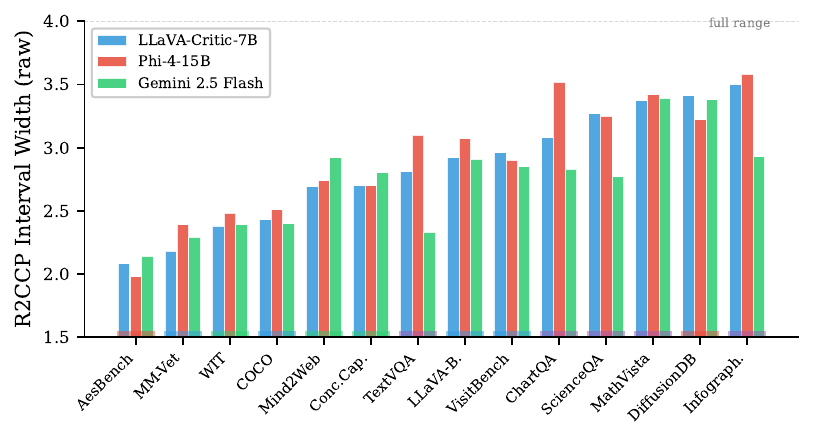}
\caption{R2CCP interval width across 14 task categories for all three judges. Task-dependent uncertainty is consistent: aesthetics tasks yield tight intervals while vision-heavy tasks produce wide intervals regardless of the judge. The ordering is largely preserved across judges (Spearman $\rho = 0.82$--$0.93$).\vspace{-15pt}}
\label{fig:task_width}
\end{figure}

Table~\ref{tab:bias2} shows that VLM judges exhibit structured and systematic errors. Low-quality responses (GT=1) are consistently overscored, often assigned scores near 3, while high-quality responses are underscored, leading to compression toward the middle of the scale. This pattern is consistent across all three judges. These structured errors explain the observed uncertainty behavior. Large and inconsistent deviations increase nonconformity scores, leading to wider intervals, particularly for vision-heavy tasks. In contrast, tasks with more consistent predictions, even if biased, produce narrower intervals. This shows that interval width reflects not only accuracy but the distribution and magnitude of errors.

\begin{table}[t]
\centering
\caption{Judge bias by ground truth score level. All judges overscore bad answers (GT=1,2) and underscore excellent ones (GT=5). Gemini has the lowest bias for bad answers.}
\label{tab:bias2}
\small
\begin{tabular}{lrrr}
\toprule
\textbf{GT Score} & \textbf{LLaVA-Critic} & \textbf{Phi-4} & \textbf{Gemini} \\
\midrule
GT$=$1 (N$=$641) & $+1.73$ & $+1.98$ & $\mathbf{+0.89}$ \\
GT$=$2 (N$=$707) & $+1.21$ & $+1.13$ & $\mathbf{+0.43}$ \\
GT$=$3 (N$=$1263) & $+0.79$ & $+0.34$ & $\mathbf{+0.23}$ \\
GT$=$4 (N$=$1791) & $+0.10$ & $-0.35$ & $\mathbf{-0.15}$ \\
GT$=$5 (N$=$1314) & $-0.74$ & $-1.04$ & $\mathbf{-0.87}$ \\
\midrule
\textbf{Overall} & $+0.38$ & $+0.08$ & $\mathbf{-0.05}$ \\
\bottomrule\vspace{-15pt}
\end{tabular}
\end{table}

\subsection{Task-dependent uncertainty}
\label{sec:task_dependent}

Table~\ref{tab:per_dataset} reveals strong task dependence in judge uncertainty, with prediction interval width varying substantially across categories. AesBench (aesthetics assessment) yields the narrowest intervals (2.08), while InfographicsVQA produces the widest (3.50), a 68\% relative difference on a 4-point scale. Equivalently, intervals cover 52\% of the score range for AesBench versus 88\% for InfographicsVQA. This variation provides a quantitative reliability map, showing that the informativeness of VLM-based evaluation depends strongly on task.

Aggregating by meta-category (Table~\ref{tab:category}) reveals a clear structure. Vision-heavy tasks produce substantially wider intervals than Knowledge/Web tasks (3.21 vs.\ 2.59), reflecting the difficulty of structured visual reasoning such as charts, diagrams, and infographics. Notably, these tasks also exhibit higher correlation, indicating that judges can rank responses but struggle to assign precise absolute scores, a pattern we analyze through ranking-scoring decoupling in \S\ref{sec:decoupling}. This pattern is consistent across all three judges (Figure~\ref{fig:task_width}). AesBench consistently yields the narrowest intervals and InfographicsVQA the widest. The Spearman rank correlation of per-dataset widths is 0.93 between LLaVA-Critic and Phi-4, and 0.82 between LLaVA-Critic and Gemini, indicating that interval width is primarily determined by the task rather than the specific model.

Overall, interval width reflects the combined effect of task difficulty, score distribution, and visual diversity. Tasks with higher accuracy and more consistent structure yield narrower intervals, while those requiring complex visual reasoning produce broader uncertainty. The observed task-dependent variability can be exploited by group-conditional conformal methods. Using Mondrian conformal prediction with task difficulty groups, we obtain narrower intervals for easy tasks (16.6\% reduction) while improving coverage on hard tasks, with similar overall coverage. This confirms that task-dependent uncertainty is structured and can be leveraged for more efficient calibration (Table~\ref{tab:mondrian}).

\begin{table}[t]
\centering
\caption{Mondrian CP vs.\ standard R2CCP (LLaVA-Critic-7B, boundary-adjusted, 10 seeds). Mondrian narrows easy-task intervals by 16.6\% while improving hard-task coverage.}
\label{tab:mondrian}
\small
\begin{tabular}{lcccc}
\toprule
& \multicolumn{2}{c}{\textbf{Standard R2CCP}} & \multicolumn{2}{c}{\textbf{Mondrian R2CCP}} \\
\cmidrule(lr){2-3} \cmidrule(lr){4-5}
\textbf{Group} & \textbf{Cov.} & \textbf{Width} & \textbf{Cov.} & \textbf{Width} \\
\midrule
Easy & 99.0\% & 3.55 & 97.5\% & \textbf{2.96} ($-$16.6\%) \\
Medium & 99.1\% & 3.60 & 98.7\% & \textbf{3.49} ($-$3.1\%) \\
Hard & 96.7\% & 3.63 & \textbf{98.2\%} & 3.78 ($+$4.1\%) \\
\midrule
Overall & 98.1\% & 3.60 & 98.2\% & \textbf{3.47} ($-$3.6\%) \\
\bottomrule \vspace{-10pt}
\end{tabular}
\end{table}

\begin{table}[t]
\centering
\caption{Average R2CCP interval width by task meta-category (LLaVA-Critic-7B, raw intervals). Vision-Heavy tasks produce 24\% wider intervals than Knowledge/Web tasks.}
\label{tab:category}
\small
\begin{tabular}{lrccc}
\toprule
\textbf{Category} & \textbf{N} & \textbf{Avg. Width} & \textbf{Avg. Pearson} & \textbf{Avg. $\pm$1 Acc.} \\
\midrule
Knowledge/Web & 1,195 & \textbf{2.59} & .265 & 78.3\% \\
General VQA & 1,448 & 2.62 & .304 & 80.3\% \\
Aesthetics/AI & 691 & 2.75 & .246 & 72.8\% \\
Vision-Heavy & 2,383 & 3.21 & .388 & 71.3\% \\
\bottomrule
\end{tabular}
\end{table}

\subsection{Ranking-scoring decoupling}
\label{sec:decoupling}

Table~\ref{tab:per_dataset} compares per-dataset correlation with interval width. ChartQA achieves the highest Pearson correlation among all 14 datasets ($\rho = 0.507$) while producing among the widest intervals (3.08). In contrast, WIT has low correlation ($\rho = 0.164$) but relatively narrow intervals (2.38). This ranking-scoring decoupling exposes a failure mode that standard evaluation metrics do not capture. A judge can correctly order responses while assigning unreliable absolute scores. For ChartQA, high correlation suggests strong performance, yet wide intervals indicate poor calibration of absolute scores. Notably, this failure mode is not well captured in standard evaluation metrics.

The effect arises because ranking depends only on monotonic consistency, while precise scoring requires accurate calibration of score magnitudes. On tasks such as chart understanding, the judge can distinguish relative quality but struggles to assign consistent absolute scores. This distinction has direct implications. For tasks where relative ordering is sufficient, correlation-based evaluation remains appropriate. For applications requiring reliable absolute scores, wide intervals indicate that point predictions are unreliable, and pairwise comparison is preferable. The decoupling is task-specific and cannot be inferred from aggregate metrics. The same pattern appears across models. For example, Gemini achieves high correlation with tight intervals on TextVQA ($\rho = 0.600$, width $= 2.33$), but low correlation with similarly wide intervals on Mind2Web ($\rho = 0.065$, width $= 2.92$), confirming that ranking quality and scoring precision are not inherently coupled.

\begin{figure}[t]
\centering
\includegraphics[width=0.7\linewidth]{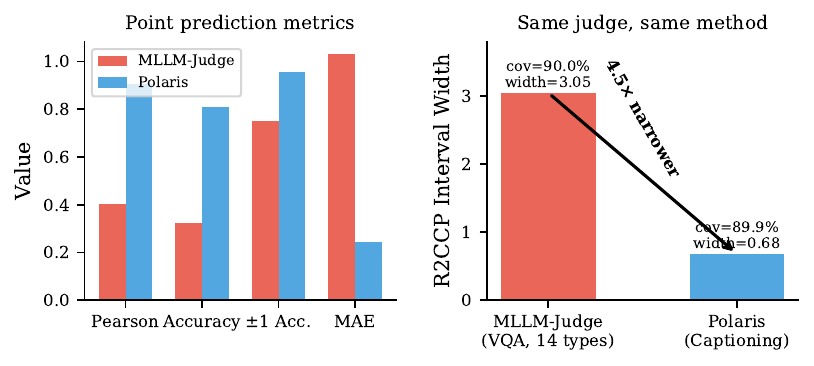}
\caption{MLLM-Judge vs.\ Polaris: same judge (LLaVA-Critic-7B), same CP method (R2CCP, $\alpha=0.10$), but $4.5\times$ narrower intervals when ground truth is clean (multi-annotator) and the task is well-defined (captioning). \vspace{-10pt}}
\label{fig:polaris}
\end{figure}

\begin{table}[t]
\centering
\caption{Same judge (LLaVA-Critic-7B) and method (R2CCP, $\alpha=0.10$). Multi-annotator GT and a well-defined task yield $4.5\times$ narrower intervals, while achieving 90\% coverage.}
\label{tab:polaris}
\small
\begin{tabular}{llcc}
\toprule
\textbf{Group} & \textbf{Metric} & \textbf{MLLM-Judge (14 tasks)} & \textbf{Polaris (Captioning)} \\
\midrule

Data & Samples & 5,717 & 8,726 \\
 & GT type & 1 annotator, integer & 3+ annotators, avg. \\
 & Task types & 14 visual tasks & 1 task \\

\midrule

Point & Pearson $\rho$ & .402 & \textbf{.906} \\
 & Accuracy & 32.2\% & \textbf{80.9\%} \\
 & $\pm 1$ Accuracy & 75.1\% & \textbf{95.4\%} \\
 & MAE & 1.031 & \textbf{0.243} \\
 & Bias & $+0.38$ & $\mathbf{-0.06}$ \\

\midrule

R2CCP & Coverage & .900 & .899 \\
 & \textbf{Width} & 3.05 (61\%) & \textbf{0.68 (14\%)} \\

\bottomrule\vspace{-20pt}
\end{tabular}
\end{table}

\subsection{Effect of ground truth quality: MLLM-Judge vs.\ Polaris}
\label{sec:polaris}

Table~\ref{tab:polaris} compares interval widths under identical conditions, using the same judge (LLaVA-Critic-7B) and conformal method (R2CCP with $\alpha = 0.10$). The results differ substantially: width is 3.05 on MLLM-Judge (61\% of the score range) and 0.68 on Polaris (14\%), a $4.5\times$ reduction, while both achieve the target 90\% coverage. This difference shows that interval width is primarily determined by the evaluation task and ground truth quality rather than the conformal method or model. Polaris is a single, well-defined captioning task with multi-annotator averaged labels, whereas MLLM-Judge spans diverse visual reasoning tasks with single-annotator scores. The resulting difference in signal quality is reflected directly in conformal intervals. The judge achieves substantially higher accuracy on Polaris (Pearson $= 0.906$ vs.\ $0.402$), leading to concentrated nonconformity scores and tighter intervals.

\textbf{Implications.}
When the task is well-defined and annotations are reliable, conformal prediction produces tight and actionable uncertainty estimates. Conversely, wide intervals on MLLM-Judge reflect genuine task difficulty and label noise rather than a limitation of the method. Notably, this advantage is judge-dependent: Phi-4-15B achieves only Pearson $= 0.128$ on Polaris despite being twice the size of LLaVA-Critic, demonstrating that model specialization for evaluation matters more than scale (Appendix~\ref{app:phi4_polaris}). Interval width should be interpreted as an indicator of evaluation reliability.

\begin{table}[t]
\centering
\caption{Multi-judge feature fusion (R2CCP, 10 seeds). Combining judges does NOT outperform the best single judge (Gemini). Logprob quality $>$ quantity.}
\label{tab:multijudge}
\small
\begin{tabular}{lrccc}
\toprule
\textbf{Configuration} & \textbf{Feat.} & \textbf{Cov.} & \textbf{Width} & \textbf{Width} \\
& & \textbf{(raw)} & \textbf{(raw)} & \textbf{(adj.)} \\
\midrule
Gemini only & 5 & .898 & \textbf{2.85} & \textbf{3.41} \\
LLaVA + Gemini & 10 & .901 & 3.02 & 3.51 \\
LLaVA only & 5 & .900 & 3.05 & 3.60 \\
Multi-Judge (all 3) & 15 & .905 & 3.14 & 3.56 \\
LLaVA + Phi-4 & 10 & .903 & 3.18 & 3.62 \\
Phi-4 only & 5 & .891 & 3.13 & 3.70 \\
\bottomrule
\end{tabular}
\end{table}

\textbf{Multi-judge feature fusion.}
We also test whether combining logprob features from multiple judges improves conformal prediction. Contrary to expectation, fusion degrades performance: combining all three judges produces wider intervals (3.14) than using the best single judge (Gemini, 2.85). This occurs because additional features from weaker judges introduce noise without sufficient signal, making the learned nonconformity mapping less effective under limited calibration data. These results suggest that feature quality is more important than feature quantity, i.e., using a single strong judge is preferable to multiple weaker ones.

\section{Conclusion}
\label{sec:conclusion}

We show that VLM judge reliability is governed primarily by task structure and data quality rather than the model or method. Uncertainty varies strongly across tasks and is consistent across judges. We identify a ranking-scoring decoupling where judges can rank responses correctly while assigning unreliable absolute scores, a failure mode not captured by standard metrics, and show that interval width is driven by task difficulty and annotation quality, with a $4.5\times$ reduction under clean, well-defined conditions. These findings lead to a simple operational guideline: narrow conformal intervals indicate reliable evaluation suitable for absolute scoring, while wide intervals signal ambiguity or noise where relative ranking or pairwise comparison is more appropriate. Conformal prediction thus provides a practical, uncertainty-aware framework for deploying VLM judges, with interval width serving as a direct indicator of evaluation reliability.

\section*{Reproducibility statement}
All experiments use publicly available models (LLaVA-Critic-7B, Phi-4-reasoning-vision-15B) and datasets (MLLM-as-a-Judge, Polaris). Gemini 2.5 Flash is accessed via the Google Vertex AI API. We will release our complete code, including inference scripts, feature extraction pipeline, conformal prediction runners, and all result CSVs upon acceptance. Results are reported across 10 random seeds with mean $\pm$ standard deviation. All conformal prediction methods use publicly available implementations (R2CCP from the authors' wheel, MAPIE v0.8.6 for CQR variants, custom implementations following published algorithms for CHR, LVD, Boosted methods, and OrdinalAPS). Hardware: 2$\times$ RTX 6000 Ada GPUs (48GB each) for open-source model inference.

\section*{Limitations}
On the MLLM-Judge benchmark, 68.8\% of boundary-adjusted intervals span more than 75\% of the rating scale, providing limited discrimination for individual evaluations. This reflects the genuine difficulty of the task and noisy single-annotator ground truth rather than a method limitation---the same approach yields decisive intervals on Polaris (width 0.68). Conditional coverage for the lowest-quality responses (GT$=$1) drops to 88.9\%, below the 90\% target, due to systematic judge overscoring of poor answers. Our score-token extraction heuristic (backward scan for the final digit) was manually verified on 100 samples per judge with 100\% accuracy, but may be less reliable on other models with different output formats. Results use a single $\alpha = 0.10$; behavior at stricter significance levels (e.g., $\alpha = 0.05$) remains unexplored. Conformal prediction requires a human-annotated calibration set, which partially offsets the cost savings of automated evaluation; however, Appendix~\ref{app:split} shows coverage is robust down to 30\% calibration splits (${\sim}1{,}700$ samples). Finally, our evaluation is specific to MLLM-Judge's 14 visual task categories and the Polaris captioning benchmark; generalization to other domains (e.g., medical imaging, document analysis) requires further validation.

\section*{Ethics statement}
Our work analyzes the reliability of automated VLM evaluators and does not introduce new evaluation capabilities---it quantifies the uncertainty of existing ones. We note several ethical considerations: (1) Conformal prediction guarantees are marginal (averaged over the calibration/test split) rather than conditional, and should not be over-interpreted as individual-instance confidence. (2) Biased human annotations in the calibration set propagate to biased prediction intervals---conformal prediction cannot correct for systematic annotator bias. (3) The positive bias we observe (judges overscoring bad answers) may mask quality problems in deployed AI systems if interval information is ignored. We encourage practitioners to use conformal intervals as a complement to, not a replacement for, careful human evaluation in high-stakes settings.


\appendix

\section*{Appendix Overview}

This appendix provides the technical and empirical details supporting the main paper. It is organized as follows. Section~\ref{app:methods} describes the conformal prediction methods, the score-token feature extraction pipeline, and inference settings for each judge. Section~\ref{app:prompts} reproduces the exact prompts used for both benchmarks in standardized prompt boxes. Section~\ref{app:hardware_software} reports hardware, software environments, and reproducibility-relevant settings. Section~\ref{app:datasets_full} describes the two benchmarks in detail, including filtering, score distributions, and license information. Section~\ref{app:core_results} contains additional cross-judge and per-dataset results. Section~\ref{app:diag} reports diagnostics: error-bin coverage, conditional coverage, judge bias, ranking-scoring decoupling, and midpoint analysis. Section~\ref{app:ablations_full} reports ablations on Mondrian conformal prediction, the naive baseline, chain-of-thought prompting, seeds, and the calibration/test split ratio. Section~\ref{app:polaris_full} contains the full Polaris analysis, and Section~\ref{app:multijudge_full} the multi-judge feature fusion study.

\section{Methods and Implementation}
\label{app:methods}

\subsection{Conformal prediction methods}
\label{sec:appendix_cp_methods}

We evaluate eight conformal prediction methods spanning regression-based and classification-based formulations, following the protocol of \citet{sheng2025analyzing}. All methods receive the same 5-dimensional score-token logprob feature vector (Eq.~\ref{eq:features}) and are calibrated against the same human ground-truth scores under the same 50/50 calibration/test split. The only difference between methods is the choice of nonconformity score and how the calibration quantile is computed.

\paragraph{Naive Split CP.}
The simplest baseline. Given a held-out predictor $\hat{y} = f(\mathbf{x})$ trained on the calibration set, the nonconformity score is the absolute residual $s_i = |y_i - \hat{y}_i|$. The $(1-\alpha)$-quantile $\hat{q}$ of $\{s_i\}$ defines the symmetric interval $[\hat{y} - \hat{q},\ \hat{y} + \hat{q}]$. This method ignores feature-conditional heterogeneity in the residual distribution.

\paragraph{Conformalized Quantile Regression (CQR)~\citep{romano2019conformalized}.}
Trains a quantile regressor for the lower and upper quantiles $\alpha/2$ and $1-\alpha/2$, then applies a symmetric conformal correction. In our setting CQR collapses to constant near-full-range intervals because the score-token logprobs are weak quantile predictors of integer human ratings; the calibration step inflates the resulting interval to recover marginal coverage.

\paragraph{Asymmetric CQR~\citep{sesia2020comparison}.}
Calibrates lower and upper quantile errors separately, allowing asymmetric corrections. In our setting it produces effectively the same $[1, 5]$ intervals as CQR after boundary adjustment.

\paragraph{Conformal Histogram Regression (CHR)~\citep{sesia2021conformal}.}
Estimates the conditional distribution $P(y \mid \mathbf{x})$ as a histogram and uses density-based nonconformity scores. Performs well in our setting (width 2.97, slight under-coverage), demonstrating that distribution-modeling methods are more robust to weak features than direct quantile regression.

\paragraph{Locally Variance-adjusted (LVD)~\citep{lin2021locally}.}
Modulates the symmetric interval width by a locally-estimated conditional standard deviation, producing wider intervals where residuals are more variable. Achieves near-target coverage but moderate width (3.21).

\paragraph{Boosted CQR and Boosted LCP~\citep{xie2024boosted}.}
Use gradient boosting to refine quantile estimates. Boosted LCP is one of our top-3 methods (width 3.02, coverage 86.3\%), while Boosted CQR over-covers due to inflated quantile estimates.

\paragraph{R2CCP~\citep{guha2024conformal}.}
Recasts continuous regression as classification over a fine grid of score values, learning the full conditional distribution $P(y \mid \mathbf{x})$ via a small neural network. Conformalization uses the negative log-density at the true label as the nonconformity score. R2CCP is our default method because it (i) hits the 90\% coverage target exactly, (ii) produces the narrowest intervals among well-calibrated methods after boundary adjustment, and (iii) is robust to weak features by modeling the full distribution rather than only the tails.

\paragraph{Ordinal Adaptive Prediction Sets (OrdinalAPS)~\citep{lu2022fair}.}
A classification-based method that constructs contiguous ordinal prediction sets. In practice it always returns the full set $\{1,2,3,4,5\}$ on our data because the integer ground-truth distribution is not separable enough at $\alpha = 0.10$.

\subsection{Score-token feature extraction}
\label{app:feature_extraction}

For each evaluation instance we extract the log-probabilities of the rating tokens at the model's final score position, yielding the 5-dimensional feature vector
\[
\mathbf{x} = [\log P(\text{``1''}), \log P(\text{``2''}), \log P(\text{``3''}), \log P(\text{``4''}), \log P(\text{``5''})].
\]
Identifying the correct score position is non-trivial because chain-of-thought outputs may contain multiple intermediate digit tokens (e.g., ``Step 2'', ``5 charts''). We use a three-stage heuristic, applied in order:

\begin{enumerate}[leftmargin=*,itemsep=2pt,topsep=2pt]
\item \textbf{Anchored format.} Search for the explicit anchor sequence \texttt{Score:} (or \texttt{Score: }) and use the first digit token in $\{1,\dots,5\}$ that follows it. This succeeds for $\geq 95\%$ of LLaVA-Critic-7B and Gemini outputs.
\item \textbf{Keyword anchor.} If the anchored format is absent, scan for the keywords \texttt{score} or \texttt{rating} (case-insensitive) followed within a small window by a rating digit.
\item \textbf{Backward scan fallback.} If neither is found, scan backwards from the end of the generated sequence and use the last token equal to a digit in $\{1,\dots,5\}$. This handles malformed outputs and is required for $\sim 1\%$ of Phi-4 reasoning traces.
\end{enumerate}

We manually verified the extraction on 100 randomly sampled outputs per judge ($300$ total) and observed 100\% extraction accuracy. Tokenization caveats: LLaVA-Critic and Phi-4 produce digits as plain tokens (\texttt{"1"} through \texttt{"5"}), but some tokenizers add leading whitespace or sentence-piece markers (e.g., \texttt{"\,1"}, \texttt{"\_1"}). Our extractor normalizes by stripping leading whitespace and known sentence-piece markers before matching.

\paragraph{Handling missing logprobs.}
Gemini 2.5 Flash returns only the top-$20$ logprobs per token via Vertex AI. When a rating token is not in the top-$20$ at the score position, we assign $\log P(\text{token}) = \log(10^{-5}) = -11.5$ as a numerical floor. This affects $< 0.5\%$ of Gemini samples and has no measurable effect on conformal coverage or width (verified by sensitivity analysis on the floor value in $\{-9, -11.5, -15\}$). For LLaVA-Critic and Phi-4, full vocabulary logprobs are available and no flooring is required.

\paragraph{NaN handling.}
A small fraction ($< 0.1\%$) of judge outputs exhibit degenerate repetition that produces NaN logprobs after \texttt{log\_softmax}. We replace any NaN with the floor value $-100.0$ and flag the sample; results are stable to inclusion or exclusion of these samples.

\subsection{Inference settings per judge}
\label{app:judge_settings}

\paragraph{LLaVA-Critic-7B} (\texttt{lmms-lab/llava-critic-7b}). Loaded with \texttt{LlavaQwenForCausalLM} from the \texttt{llava} package, fp16, \texttt{attn\_implementation="sdpa"}. We found that the \texttt{eager} attention implementation corrupts the model's Sliding Window Attention and produces garbage outputs; SDPA is required. We use greedy decoding (\texttt{do\_sample=False}), \texttt{max\_new\_tokens=512}, \texttt{output\_scores=True}, \texttt{return\_dict\_in\_generate=True}, with \texttt{top\_k=10} kept only for the saved logprob top-$k$.

\paragraph{Phi-4-reasoning-vision-15B} (\texttt{microsoft/Phi-4-reasoning}). Loaded in \texttt{phi4\_env} (transformers $\geq 4.57.1$ for \texttt{Siglip2VisionModel} support), fp16, SDPA attention. We use greedy decoding with \texttt{max\_new\_tokens=2048} to accommodate extended reasoning chains. We additionally evaluate a no-think variant where we set \texttt{enable\_thinking=False} (Polaris only).

\paragraph{Gemini 2.5 Flash} (Google Vertex AI, \texttt{gemini-2.5-flash}). Accessed via the \texttt{litellm} interface with \texttt{response\_logprobs=True} and \texttt{top\_logprobs=20}. We use temperature $0$ and the default safety settings. Authentication uses Application Default Credentials with a Vertex AI quota project.

\subsection{R2CCP training and quantile computation}
\label{app:r2ccp_train}

R2CCP discretizes the score range $[1, 5]$ into a fine grid (we use $K_{\text{grid}} = 41$ points spanning $[0.5, 5.5]$ at $0.125$ resolution). A small MLP with two hidden layers of $64$ and $32$ units maps $\mathbf{x}$ to a softmax distribution over the grid, trained with cross-entropy against the nearest-grid-point label on the calibration set. The nonconformity score is $s(\mathbf{x}, y) = -\log \hat{p}(y \mid \mathbf{x})$ where $\hat{p}$ is the predicted distribution. The conformal threshold is the $\lceil (n+1)(1-\alpha) \rceil / n$-th empirical quantile of calibration scores. At test time, the prediction interval is the smallest contiguous grid range $[l, u]$ whose negative log-density does not exceed the threshold. We fix the seed for the MLP initialization to $42$ but vary the calibration/test split seed across $\{0, 1, \dots, 9\}$.

\paragraph{Boundary adjustment.}
For all methods, we report both raw continuous intervals $[l, u]$ and integer-aligned intervals $[\lceil l \rceil, \lfloor u \rfloor]$. The integer-aligned form aligns endpoints with valid Likert labels. Boundary adjustment is monotone (it only expands the interval) and preserves coverage; in our discrete-GT setting it gains $+8$pp coverage on average (cf. \citealp{sheng2025analyzing}'s $+2$--$5$pp on continuous-averaged GT).

\section{Evaluation prompts}
\label{app:prompts}

We use a single, deterministic prompt template per benchmark. Both prompts are chain-of-thought style and explicitly anchor the final score with the literal string \texttt{Score: X}. Variables in the template are denoted with curly braces (e.g., \texttt{\{question\}}) and are substituted at runtime. We never alter the prompt across judges, datasets, or seeds. The prompts are reproduced verbatim below.

\subsection{MLLM-as-a-Judge prompt}
\label{app:prompt_mllm}

\begin{promptbox}[Prompt 1: MLLM-as-a-Judge (CoT scoring)]
\small
\textbf{System role:} Please serve as an unbiased judge in assessing the quality of the response from an AI assistant regarding the user's instruction and the provided image.\\[2pt]

\textbf{Evaluation steps:} Please examine the provided image attentively. Begin by conducting a comprehensive analysis of the figure provided. Then, utilize the insights from your analysis to critically evaluate the response. Finally, based on your figure analysis and response evaluation, form a well-reasoned judgement.\\[2pt]

\textbf{Scoring rubric:}
\begin{itemize}[leftmargin=14pt,itemsep=1pt,topsep=2pt]
\item \textbf{Poor (1):} The response significantly deviates from the user's instruction and fails to address the query effectively. It shows a lack of relevance, accuracy, and comprehensiveness. Creativity and granularity are absent or poorly executed.
\item \textbf{Fair (2):} The response addresses the user's instruction partially, with evident shortcomings in relevance, accuracy, or comprehensiveness. It lacks depth in creativity and granularity.
\item \textbf{Average (3):} The response adequately addresses the user's instruction, showing a fair level of relevance, accuracy, and comprehensiveness. It reflects a basic level of creativity and granularity but may lack sophistication.
\item \textbf{Good (4):} The response is well-aligned with the user's instruction, demonstrating a high degree of relevance, accuracy, and comprehensiveness. It shows creativity and nuanced understanding.
\item \textbf{Excellent (5):} The response perfectly adheres to the user's instruction, excelling in relevance, accuracy, comprehensiveness, creativity, and granularity.
\end{itemize}

\textbf{Notice:} In your evaluation, weigh factors such as relevance, accuracy, comprehensiveness, creativity, and the granularity of the response. Do not allow the length of the response to influence your evaluation. Be as objective as possible.\\[2pt]

\textbf{Input:}\\
{[}The Start of User Instruction{]} \texttt{\{question\}} {[}The End of User Instruction{]}\\
{[}The Start of Assistant's Answer{]} \texttt{\{actor\_answer\}} {[}The End of Assistant's Answer{]}\\[2pt]

\textbf{Output format:} First, provide your analysis of the image and the response. Then, at the very end of your response, provide your final score in exactly this format on its own line: \texttt{``Score: X''} where \texttt{X} is a single integer from 1 to 5.
\end{promptbox}

\paragraph{Variables.} \texttt{\{question\}} is the user instruction (text), \texttt{\{actor\_answer\}} is the candidate response from a held-out actor model. The image is supplied through the model's native vision input channel and is not embedded in the text prompt. \texttt{\{actor\_answer\}} is taken from the original MLLM-as-a-Judge release (we do not regenerate actor responses) so that human ground-truth scores apply.

\paragraph{Expected output.} The judge produces a free-form chain-of-thought analysis followed by a single line of the form \texttt{Score: X}. We extract \texttt{X} and the corresponding score-token logprobs as described in \S\ref{app:feature_extraction}. Outputs that do not conform to this format are routed through the backward-scan fallback (affects $< 1\%$ of samples).

\subsection{Polaris captioning prompt}
\label{app:prompt_polaris}

\begin{promptbox}[Prompt 2: Polaris (caption-quality scoring)]
\small
\textbf{System role:} Please serve as an unbiased judge in assessing the quality of an image caption generated by an AI model.\\[2pt]

\textbf{Evaluation steps:} Please examine the provided image attentively. Then evaluate how well the generated caption describes the image content.\\[2pt]

\textbf{Scoring rubric:}
\begin{itemize}[leftmargin=14pt,itemsep=1pt,topsep=2pt]
\item \textbf{Poor (1):} The caption is completely inaccurate, irrelevant, or fails to describe the image content. Major factual errors or hallucinations are present.
\item \textbf{Fair (2):} The caption partially describes the image but has significant inaccuracies, missing key elements, or includes incorrect details.
\item \textbf{Average (3):} The caption adequately describes the main content of the image with fair accuracy, but may miss some details or include minor inaccuracies.
\item \textbf{Good (4):} The caption accurately describes the image content with good detail and relevance, with only minor omissions.
\item \textbf{Excellent (5):} The caption perfectly describes the image content with high accuracy, comprehensive detail, and excellent relevance to what is shown.
\end{itemize}

\textbf{Input:}\\
{[}The Start of Caption{]} \texttt{\{caption\}} {[}The End of Caption{]}\\[2pt]

\textbf{Output format:} Provide your analysis of how well the caption matches the image, then provide your final score in exactly this format on its own line: \texttt{``Score: X''} where \texttt{X} is a single integer from 1 to 5.
\end{promptbox}

\paragraph{Variables.} \texttt{\{caption\}} is the candidate caption produced by an actor model. The image is supplied via the model's native vision input. We use Polaris's averaged human caption-quality score as ground truth, mapped from its native scale to a 1--5 Likert range by binning at integer boundaries (cf. \S\ref{app:datasets_full}).

\subsection{Why these prompts}

The MLLM-as-a-Judge prompt closely follows the ``COT figure'' setting from \citet{chen2024mllmasajudge} so that our point-prediction baseline is directly comparable to the original benchmark numbers. The Polaris prompt is a caption-quality variant of the same template, simplified to a single content dimension (caption-vs-image accuracy) since Polaris does not break out separate sub-dimensions. We deliberately keep the prompts minimal and identical across judges to isolate the effect of the underlying model and feature representation; differences in calibration, bias, and width across judges therefore cannot be attributed to prompt variation.

\section{Hardware, software, and reproducibility}
\label{app:hardware_software}

\paragraph{Hardware.} All open-source inference and conformal calibration runs use 2$\times$ NVIDIA RTX 6000 Ada GPUs (48~GB VRAM each, 94.8~GB total). LLaVA-Critic-7B and Phi-4-reasoning-vision-15B both fit comfortably in fp16 across the two GPUs via Hugging Face \texttt{device\_map="auto"}. Gemini 2.5 Flash inference is API-based and incurs no local GPU cost.

\paragraph{Software.} We use two conda environments to handle differing transformers requirements. The main environment (\texttt{env\_py311}, Python 3.11) uses \texttt{transformers==4.47.0}, \texttt{torch==2.5.1}, \texttt{mapie==0.8.6}, and the R2CCP wheel installed with \texttt{pip install R2CCP-0.0.8-py3-none-any.whl --no-deps}. MAPIE must be pinned to 0.8.6: the v1.x API renames \texttt{MapieQuantileRegressor} and breaks the CQR variants. The second environment (\texttt{phi4\_env}) uses \texttt{transformers==4.57.1} (required for \texttt{Siglip2VisionModel} not present in 4.47). Gemini access uses \texttt{litellm} configured with Vertex AI Application Default Credentials.

\paragraph{Determinism.} VLM inference is greedy and deterministic. Conformal calibration involves randomness only in the calibration/test split, which we control through a single seed in $\{0,\dots,9\}$. R2CCP's MLP is initialized from a fixed seed (42) and trained to a fixed number of epochs. We confirm in Appendix~\ref{app:seeds} that means and standard deviations stabilize within 5--10 seeds.

\paragraph{Wall-clock budget.} A full pipeline (judge inference $+$ feature extraction $+$ all 8 conformal methods, 10 seeds, on MLLM-as-a-Judge) takes approximately 6 hours for LLaVA-Critic, 14 hours for Phi-4 (slower due to long reasoning chains), and 3 hours for Gemini (rate-limited API). Conformal calibration alone, given pre-extracted features, runs in under 10 minutes per judge.

\section{Datasets}
\label{app:datasets_full}

\subsection{MLLM-as-a-Judge}

We use the score evaluation split of the MLLM-as-a-Judge benchmark~\citep{chen2024mllmasajudge}, downloaded from the official repository (\texttt{Dongping-Chen/MLLM-Judge}). Each instance contains an image, a user instruction, an actor-model-generated answer, and a single human annotator's score on a 1--5 Likert scale. We retain $5{,}717$ samples after removing instances with missing images or malformed scores. The dataset spans 14 task categories grouped into four meta-categories:

\begin{itemize}[leftmargin=14pt,itemsep=1pt,topsep=2pt]
\item \textbf{Vision-heavy:} ChartQA, ScienceQA, MathVista, InfographicsVQA, TextVQA. Tasks requiring fine-grained visual parsing or multi-step visual-numerical reasoning.
\item \textbf{Knowledge/Web:} WIT, Mind2Web, Conceptual Captions. Tasks dominated by textual world knowledge with image as context.
\item \textbf{General VQA:} COCO, MM-Vet, LLaVA-Bench, VisitBench. Open-ended visual question answering on natural images.
\item \textbf{Aesthetics/AI:} AesBench, DiffusionDB. Subjective image-quality assessment and AI-generated image scoring.
\end{itemize}

The score distribution is right-skewed: $\sim 54\%$ of samples have GT $\in \{4, 5\}$, only $\sim 11\%$ have GT $= 1$. This skew affects both calibration of the lowest-quality stratum and the magnitude of judge bias (\S\ref{app:bias}).

\subsection{Polaris}

Polaris~\citep{wada2024polos} is an image-captioning evaluation benchmark with $8{,}726$ image--caption pairs scored by multiple human annotators. We use the released averaged human score as ground truth and bin it to a 1--5 Likert range by integer boundaries: a score of $[1.0, 1.5)$ maps to 1, $[1.5, 2.5)$ to 2, and so on. The Polaris score distribution is more balanced and the multi-annotator averaging reduces label noise relative to MLLM-as-a-Judge's single-annotator setup. This makes Polaris ideal as a clean, well-defined contrast point for assessing the role of label quality (\S\ref{sec:polaris}).

\subsection{Licensing and access}

MLLM-as-a-Judge is released under the CC-BY-NC license. Polaris is released under CC-BY-4.0. LLaVA-Critic-7B is released under Apache-2.0 (\texttt{lmms-lab/llava-critic-7b} on Hugging Face). Phi-4-reasoning is released under the MIT license (\texttt{microsoft/Phi-4-reasoning} on Hugging Face). Gemini 2.5 Flash is accessed via the Google Vertex AI commercial API. All datasets and open-source models are downloaded once and cached locally; no derivative datasets are created.

\section{Core results}
\label{app:core_results}

\subsection{Cross-judge comparison across all methods}
\label{app:all_methods}

Table~\ref{tab:all_methods_all_judges} extends the main-text Table~\ref{tab:cross_judge} by reporting all five top conformal methods on each of the three judges. R2CCP and CHR are consistently the best two methods on every judge, with Gemini producing the narrowest intervals across all five methods. The CQR row is included as a sanity check: on every judge it collapses to near-full-range coverage of $\sim 0.996$ with width $\sim 3.95$, confirming that this failure is intrinsic to the method-feature combination rather than a property of any single judge.

\begin{table}[t]
\centering
\caption{Top conformal prediction methods across all three judges (raw coverage and width, $\alpha = 0.10$, 10 seeds). Bold indicates narrowest width per method.}
\label{tab:all_methods_all_judges}
\small
\begin{tabular}{lcccccc}
\toprule
& \multicolumn{2}{c}{\textbf{LLaVA-Critic}} & \multicolumn{2}{c}{\textbf{Phi-4}} & \multicolumn{2}{c}{\textbf{Gemini}} \\
\cmidrule(lr){2-3} \cmidrule(lr){4-5} \cmidrule(lr){6-7}
\textbf{Method} & Cov. & Width & Cov. & Width & Cov. & Width \\
\midrule
R2CCP & .900 & 3.05 & .891 & 3.13 & .898 & \textbf{2.85} \\
CHR & .880 & 2.97 & .896 & 3.16 & .880 & \textbf{2.80} \\
Boosted LCP & .863 & 3.02 & .875 & 3.24 & .863 & \textbf{2.88} \\
Naive Split CP & .895 & 3.23 & .897 & 3.30 & .893 & \textbf{3.08} \\
CQR & .996 & 3.95 & .997 & 3.98 & .996 & 3.95 \\
\bottomrule
\end{tabular}
\end{table}

\subsection{Per-dataset width across judges}

Table~\ref{tab:all_judges_per_dataset} reports per-dataset R2CCP raw width for all three judges. Two patterns emerge. First, the relative ordering of datasets by width is highly consistent across judges (Spearman $\rho = 0.93$ between LLaVA-Critic and Phi-4, $\rho = 0.82$ between LLaVA-Critic and Gemini), supporting the main claim that interval width is determined primarily by task properties rather than judge identity. Second, Gemini achieves the narrowest interval on $10$ of $14$ datasets, with the largest gains concentrated on Vision-Heavy categories (TextVQA, ChartQA, ScienceQA, InfographicsVQA), suggesting that Gemini's logprob distributions retain stronger discriminative signal on tasks requiring fine-grained visual reasoning.

\begin{table}[t]
\centering
\caption{Per-dataset R2CCP raw width for all three judges (10 seeds). Bold indicates narrowest width for each dataset. Gemini is best on 10 of 14 datasets.}
\label{tab:all_judges_per_dataset}
\small
\begin{tabular}{lcccc}
\toprule
\textbf{Dataset} & \textbf{LLaVA} & \textbf{Phi-4} & \textbf{Gemini} & \textbf{Cat.} \\
\midrule
AesBench & 2.08 & \textbf{1.98} & 2.14 & Aesth. \\
MM-Vet & \textbf{2.18} & 2.39 & 2.29 & Gen. \\
WIT & \textbf{2.38} & 2.48 & 2.39 & Know. \\
COCO & 2.43 & 2.51 & \textbf{2.40} & Gen. \\
Mind2Web & \textbf{2.69} & 2.74 & 2.92 & Know. \\
Conc.\ Cap. & 2.70 & \textbf{2.70} & 2.80 & Know. \\
TextVQA & 2.81 & 3.10 & \textbf{2.33} & Vision \\
LLaVA-B. & 2.92 & 3.07 & \textbf{2.91} & Gen. \\
VisitBench & 2.96 & 2.90 & \textbf{2.85} & Gen. \\
ChartQA & 3.08 & 3.52 & \textbf{2.83} & Vision \\
ScienceQA & 3.27 & 3.25 & \textbf{2.77} & Vision \\
MathVista & 3.37 & 3.42 & \textbf{3.39} & Vision \\
DiffusionDB & 3.41 & \textbf{3.22} & 3.38 & Aesth. \\
Infograph. & 3.50 & 3.58 & \textbf{2.93} & Vision \\
\midrule
\textbf{Overall} & 3.05 & 3.13 & \textbf{2.85} & --- \\
\bottomrule
\end{tabular}
\end{table}

\subsection{Per-dataset relaxed accuracy and MAE}
\label{app:relaxed}

Table~\ref{tab:relaxed_accuracy} reports per-dataset relaxed accuracy ($\pm 1$ tolerance) and mean absolute error for LLaVA-Critic-7B. Datasets are sorted by $\pm 1$ accuracy. The relationship between accuracy and interval width is monotone but not strict: easier tasks (AesBench, COCO, Conceptual Captions) yield both higher $\pm 1$ accuracy and narrower intervals, while harder tasks (DiffusionDB, MathVista, Infographics) exhibit both lower accuracy and wider intervals. The few violations of strict monotonicity (e.g., ChartQA has higher $\pm 1$ accuracy but wider intervals than LLaVA-Bench) are precisely the ranking-scoring decoupling cases analyzed in \S\ref{sec:decoupling}.

\begin{table}[t]
\centering
\caption{Per-dataset exact and $\pm 1$ accuracy, MAE, and bias for LLaVA-Critic-7B on MLLM-as-a-Judge.}
\label{tab:relaxed_accuracy}
\small
\begin{tabular}{lrcccc}
\toprule
\textbf{Dataset} & \textbf{N} & \textbf{Exact} & $\pm$\textbf{1} & \textbf{MAE} & \textbf{Bias} \\
\midrule
AesBench & 392 & 41.3\% & 88.3\% & 0.732 & $+0.49$ \\
COCO & 397 & 34.5\% & 84.9\% & 0.854 & $+0.04$ \\
Conc.\ Cap. & 398 & 35.4\% & 81.7\% & 0.872 & $+0.60$ \\
LLaVA-B. & 396 & 31.8\% & 81.1\% & 0.922 & $+0.55$ \\
MM-Vet & 258 & 37.2\% & 80.2\% & 0.926 & $-0.01$ \\
TextVQA & 399 & 36.3\% & 79.2\% & 0.942 & $+0.30$ \\
WIT & 399 & 33.8\% & 78.7\% & 0.912 & $+0.51$ \\
VisitBench & 397 & 31.7\% & 75.1\% & 1.018 & $+0.44$ \\
Mind2Web & 398 & 29.6\% & 74.4\% & 1.008 & $+0.19$ \\
ChartQA & 400 & 30.2\% & 73.2\% & 1.085 & $+0.31$ \\
ScienceQA & 396 & 26.5\% & 69.7\% & 1.200 & $+0.30$ \\
Infograph. & 398 & 25.1\% & 69.3\% & 1.191 & $+0.18$ \\
MathVista & 789 & 32.3\% & 65.1\% & 1.235 & $+0.33$ \\
DiffusionDB & 299 & 23.7\% & 57.2\% & 1.385 & $+1.23$ \\
\bottomrule
\end{tabular}
\end{table}

\subsection{Confusion matrices for all three judges}
\label{app:confusion}

Table~\ref{tab:confusion} reports row-normalized confusion matrices for the three judges on MLLM-as-a-Judge. Each judge exhibits a distinct error structure. \textbf{LLaVA-Critic} shows broad dispersion across the entire prediction range, suggesting it uses the full Likert scale but is noisily calibrated. \textbf{Phi-4} concentrates strongly around score 4 regardless of ground truth, consistent with a strong central-tendency bias likely inherited from the model's instruction-tuning distribution. \textbf{Gemini} polarizes between scores 1 and 5, with relatively low usage of the middle scores 2--3; this matches the high-confidence behavior reflected in its softmax overconfidence ($83\%$ of samples have softmax max $> 0.99$) but does not prevent it from achieving the highest correlation among the three judges.

\begin{table}[t]
\centering
\caption{Row-normalized confusion matrices (\%, ground-truth rows, prediction columns) for the three judges on MLLM-as-a-Judge. \textit{LLaVA-Critic disperses, Phi-4 gravitates to 4, Gemini polarizes to 1 and 5.}}
\label{tab:confusion}
\small
\begin{tabular}{l|ccccc|c}
\toprule
& \textbf{P=1} & \textbf{P=2} & \textbf{P=3} & \textbf{P=4} & \textbf{P=5} & $\pm$\textbf{1} \\
\midrule
\multicolumn{7}{l}{\emph{LLaVA-Critic-7B}} \\
GT=1 & 32.6 & 12.9 & 17.0 & 23.4 & 14.0 & 45.6 \\
GT=2 & 17.0 & 11.7 & 21.9 & 32.2 & 17.1 & 50.6 \\
GT=3 & 5.2 & 7.0 & 18.6 & 42.0 & 27.2 & 67.6 \\
GT=4 & 3.3 & 3.6 & 13.4 & 38.9 & 40.8 & 93.1 \\
GT=5 & 1.9 & 2.2 & 10.7 & 38.4 & 46.7 & 85.2 \\
\midrule
\multicolumn{7}{l}{\emph{Phi-4-reasoning-15B}} \\
GT=1 & 21.6 & 18.1 & 13.8 & 33.6 & 13.0 & 39.6 \\
GT=2 & 12.8 & 18.4 & 21.5 & 37.9 & 9.4 & 52.7 \\
GT=3 & 6.4 & 13.3 & 28.1 & 44.5 & 7.7 & 85.9 \\
GT=4 & 2.4 & 10.2 & 19.5 & 56.2 & 11.8 & 87.4 \\
GT=5 & 2.5 & 4.6 & 11.4 & 56.9 & 24.6 & 81.4 \\
\midrule
\multicolumn{7}{l}{\emph{Gemini 2.5 Flash}} \\
GT=1 & 61.7 & 15.5 & 6.4 & 5.2 & 11.3 & 77.1 \\
GT=2 & 38.7 & 24.7 & 9.3 & 9.9 & 17.5 & 72.7 \\
GT=3 & 17.6 & 21.3 & 12.0 & 18.8 & 30.2 & 52.1 \\
GT=4 & 10.4 & 13.7 & 9.0 & 15.0 & 52.0 & 76.0 \\
GT=5 & 7.9 & 10.3 & 6.7 & 11.4 & 63.7 & 75.0 \\
\bottomrule
\end{tabular}
\end{table}

\section{Diagnostics}
\label{app:diag}

\subsection{Error-bin coverage}
\label{app:error_bins}

Table~\ref{tab:error_bins} stratifies test samples by the magnitude of the judge's point error $|y - \hat{y}|$ and reports both raw and boundary-adjusted coverage of the conformal interval. The result motivates the practical utility of CP: the conformal interval is precisely most useful when the judge is wrong. Boundary-adjusted coverage exceeds 99\% on the $\pm 1$ and $\pm 2$ error bins (which together account for $59\%$ of all samples) and reaches $91.5\%$ on the $\pm 3$ bin. Even on $\pm 4$ errors (where the judge is grossly wrong), boundary-adjusted coverage is $43.3\%$, meaning the interval still contains the true score nearly half the time. Aggregating, $97.8\%$ of all judge errors are ``recovered'' by the boundary-adjusted CP interval, in the sense that the true human score lies within the predicted interval.

\begin{table}[t]
\centering
\caption{Conformal coverage and width by judge-error magnitude (R2CCP, LLaVA-Critic-7B, 10 seeds). Boundary adjustment recovers $97.8\%$ of judge errors overall.}
\label{tab:error_bins}
\small
\begin{tabular}{lccccc}
\toprule
\textbf{Error} & \textbf{\% samples} & \textbf{Cov. (raw)} & \textbf{Cov. (adj.)} & \textbf{Width (raw)} & \textbf{Width (adj.)} \\
\midrule
0 (exact) & 32.6\% & 99.8\% & 100\% & 2.98 & 3.55 \\
$\pm 1$ & 43.0\% & 98.7\% & 99.9\% & 3.08 & 3.63 \\
$\pm 2$ & 16.0\% & 84.3\% & 99.4\% & 3.12 & 3.65 \\
$\pm 3$ & 6.3\% & 25.4\% & 91.5\% & 3.05 & 3.61 \\
$\pm 4$ & 2.0\% & 5.7\% & 43.3\% & 2.89 & 3.42 \\
\midrule
\multicolumn{6}{l}{\emph{Overall judge-error recovery: 97.8\%}} \\
\bottomrule
\end{tabular}
\end{table}

\subsection{Cross-judge error-bin coverage}
\label{app:cross_judge_errors}

Table~\ref{tab:error_bins_all} extends the error-bin analysis to all three judges. The recovery pattern is consistent: boundary-adjusted coverage exceeds $99\%$ on the $\pm 1$ and $\pm 2$ bins for every judge. Gemini achieves the highest coverage on larger errors ($\pm 3$: $96.2\%$ vs.\ $91.5\%$ for LLaVA), reflecting the fact that its narrower but better-centered intervals more often span the true score even when it is several Likert points away from the point prediction.

\begin{table}[t]
\centering
\caption{Boundary-adjusted CP coverage by judge-error magnitude across all three judges (R2CCP, 10 seeds).}
\label{tab:error_bins_all}
\small
\begin{tabular}{lccc}
\toprule
\textbf{Error} & \textbf{LLaVA} & \textbf{Phi-4} & \textbf{Gemini} \\
\midrule
Exact & 100\% & 99.8\% & 99.8\% \\
$\pm 1$ & 99.9\% & 99.9\% & 99.8\% \\
$\pm 2$ & 99.4\% & 99.5\% & 99.0\% \\
$\pm 3$ & 91.5\% & 86.3\% & \textbf{96.2\%} \\
$\pm 4$ & 43.3\% & 67.4\% & \textbf{68.8\%} \\
\bottomrule
\end{tabular}
\end{table}

\subsection{Conditional coverage by ground truth}
\label{app:conditional}

Table~\ref{tab:conditional} reports coverage stratified by ground-truth Likert level. Coverage is below the $90\%$ marginal target only for GT $= 1$ ($88.9\%$ adjusted), where systematic judge overscoring (mean bias $+1.67$) places the point predictor far from the true low score, requiring wider intervals to recover. For GT $\in \{2, 3, 4\}$ the boundary-adjusted coverage is essentially $100\%$. This is a known limitation of marginal CP: marginal coverage does not imply conditional coverage, and our reported $\geq 90\%$ guarantee should be interpreted as an averaged-over-the-test-distribution guarantee.

\begin{table}[t]
\centering
\caption{Conditional coverage by ground-truth score level (R2CCP, LLaVA-Critic-7B). GT $=1$ under-covers due to systematic overscoring of poor responses.}
\label{tab:conditional}
\small
\begin{tabular}{lcccccc}
\toprule
\textbf{GT} & \textbf{N} & \textbf{Cov. (raw)} & \textbf{Cov. (adj.)} & \textbf{Width (raw)} & \textbf{Width (adj.)} & \textbf{Bias} \\
\midrule
1 & 334 & 53.0\% & 88.9\% & 3.14 & 3.75 & $+1.67$ \\
2 & 360 & 86.9\% & 100\% & 3.18 & 3.79 & $+1.13$ \\
3 & 637 & 99.7\% & 100\% & 3.09 & 3.75 & $+0.77$ \\
4 & 880 & 98.2\% & 100\% & 2.99 & 3.65 & $+0.11$ \\
5 & 648 & 92.4\% & 99.2\% & 2.92 & 3.58 & $-0.80$ \\
\bottomrule
\end{tabular}
\end{table}

\subsection{Judge bias by ground-truth score}
\label{app:bias}

Table~\ref{tab:bias} reports mean signed error (judge minus human) stratified by ground-truth level for each judge. All three judges exhibit a structured ``compression'' pattern: low-quality responses (GT $\in \{1, 2\}$) are systematically overscored, high-quality responses (GT $= 5$) are systematically underscored, and the mid-range (GT $= 3$) receives the smallest absolute bias. This pattern is most severe for LLaVA-Critic and Phi-4, and least severe for Gemini, which on bad answers (GT $= 1$) is roughly half as biased as the open-source judges.

\begin{table}[t]
\centering
\caption{Mean signed error (judge $-$ human) by ground-truth score level. All judges overscore low quality and underscore high quality; Gemini is least biased on poor answers.}
\label{tab:bias}
\small
\begin{tabular}{lccc}
\toprule
\textbf{GT Score} & \textbf{LLaVA-Critic} & \textbf{Phi-4} & \textbf{Gemini} \\
\midrule
1 & $+1.73$ & $+1.98$ & $\mathbf{+0.89}$ \\
2 & $+1.21$ & $+1.13$ & $\mathbf{+0.43}$ \\
3 & $+0.79$ & $+0.34$ & $\mathbf{+0.23}$ \\
4 & $+0.10$ & $-0.35$ & $\mathbf{-0.15}$ \\
5 & $-0.74$ & $-1.04$ & $\mathbf{-0.87}$ \\
\midrule
\textbf{Overall} & $+0.38$ & $+0.08$ & $\mathbf{-0.05}$ \\
\bottomrule
\end{tabular}
\end{table}

\subsection{Ranking-scoring gap (RSG) and decoupling figure}
\label{app:rsg}

Table~\ref{tab:rsg} reports the per-dataset Ranking-Scoring Gap (RSG) for all three judges. Recall RSG $= |\rho| - (1 - w/(K-1))$. Positive RSG means the judge is a stronger ranker than scorer (high correlation, wide intervals); negative RSG means the judge is a stronger scorer than ranker (narrow intervals despite weaker correlation). The vision-heavy tasks at the top of the table (Infographics, ChartQA, MathVista) consistently exhibit positive RSG across all three judges, while knowledge/aesthetics tasks at the bottom (WIT, MM-Vet, AesBench) consistently exhibit negative RSG. RSG is therefore a property of the task, not of the judge.

\begin{table}[t]
\centering
\caption{Per-dataset Ranking-Scoring Gap (RSG) for all three judges. Positive: strong ranker / weak scorer; Negative: weak ranker / strong scorer. Bold marks $|\text{RSG}| > 0.2$.}
\label{tab:rsg}
\small
\begin{tabular}{lcccccc}
\toprule
& \multicolumn{2}{c}{\textbf{LLaVA}} & \multicolumn{2}{c}{\textbf{Phi-4}} & \multicolumn{2}{c}{\textbf{Gemini}} \\
\cmidrule(lr){2-3} \cmidrule(lr){4-5} \cmidrule(lr){6-7}
\textbf{Dataset} & $\rho$ & RSG & $\rho$ & RSG & $\rho$ & RSG \\
\midrule
Infograph. & .411 & \textbf{+.287} & .162 & +.058 & .547 & \textbf{+.281} \\
ChartQA & .507 & \textbf{+.276} & .261 & +.141 & .494 & \textbf{+.201} \\
MathVista & .376 & \textbf{+.219} & .339 & +.194 & .414 & \textbf{+.262} \\
VisitBench & .352 & +.091 & .339 & +.064 & .524 & \textbf{+.235} \\
ScienceQA & .258 & +.075 & .304 & +.115 & .530 & \textbf{+.223} \\
TextVQA & .389 & +.092 & .213 & $-$.012 & .600 & +.183 \\
\midrule
WIT & .164 & \textbf{$-$.242} & .400 & +.021 & .294 & $-$.109 \\
MM-Vet & .260 & +.195 & .296 & $-$.108 & .233 & $-$.194 \\
AesBench & .402 & $-$.077 & .353 & $-$.153 & .256 & \textbf{$-$.208} \\
\bottomrule
\end{tabular}
\end{table}

\begin{figure}[t]
\centering
\includegraphics[width=\linewidth]{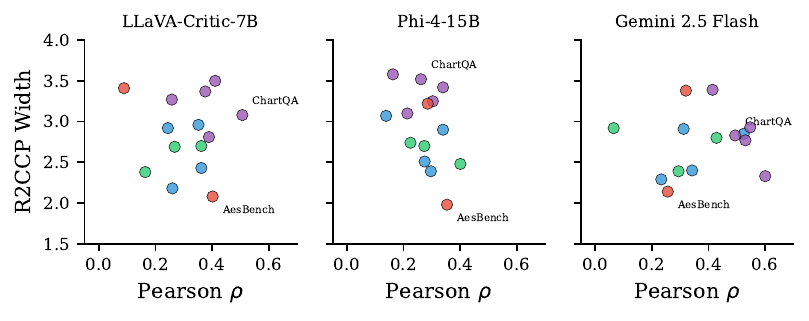}
\caption{Ranking-scoring decoupling across datasets. The horizontal axis is Pearson correlation, the vertical axis is interval width. Points in the upper-right corner (high correlation, wide intervals) are the decoupling cases: the judge ranks well but cannot assign reliable absolute scores.}
\label{fig:decoupling}
\end{figure}

\subsection{Interval informativeness and midpoint analysis}
\label{app:midpoint}

Of the boundary-adjusted intervals on MLLM-as-a-Judge, $0.5\%$ are decisive (width $\leq 1$ on the 1--5 scale), $30.7\%$ are moderately informative ($1 <$ width $\leq 3$), and $68.8\%$ are uninformative (width $> 3$). On Polaris these proportions invert: the majority of intervals are decisive due to the much higher signal quality (\S\ref{app:polaris_full}). Table~\ref{tab:midpoint} compares the raw judge score, the conformal interval midpoint, and the boundary-adjusted midpoint as point estimates. The midpoint does not consistently improve over the raw score, primarily because the wide intervals on MLLM-Judge cause the midpoint to regress toward the centre of the scale.

\begin{table}[t]
\centering
\caption{Midpoint vs.\ raw score as point estimates (LLaVA-Critic-7B, MLLM-Judge).}
\label{tab:midpoint}
\small
\begin{tabular}{lccc}
\toprule
\textbf{Metric} & \textbf{Raw} & \textbf{Midpoint} & \textbf{Adj.\ Mid} \\
\midrule
Pearson & .412 & .396 & .277 \\
Spearman & .367 & .359 & .272 \\
MAE & 1.022 & .992 & 1.056 \\
Accuracy & 32.6\% & --- & 28.1\% \\
\bottomrule
\end{tabular}
\end{table}

\section{Ablations}
\label{app:ablations_full}

\subsection{Mondrian conformal prediction}
\label{app:mondrian}

Standard CP uses a single global quantile $\hat{q}$ for all test samples. Mondrian (group-conditional) CP partitions the calibration set into disjoint groups and computes a separate quantile per group, yielding group-conditional coverage guarantees of the form $\mathbb{P}(y \in \mathcal{C}(\mathbf{x}) \mid G = g) \geq 1 - \alpha$. We define three groups by initial per-dataset R2CCP width on a held-out pilot split: \emph{easy} (width $< 2.5$: AesBench, MM-Vet, WIT, COCO), \emph{medium} ($2.5$--$3.2$: Mind2Web, Conceptual Captions, TextVQA, LLaVA-Bench, VisitBench, ChartQA), and \emph{hard} ($> 3.2$: ScienceQA, MathVista, DiffusionDB, Infographics). Group boundaries are fixed once and not retuned across seeds.

The result (Table~\ref{tab:mondrian_app}, identical to the main-text Table~\ref{tab:mondrian} reproduced here for completeness) is that Mondrian CP narrows easy-group intervals by $16.6\%$ while improving hard-group coverage by $1.5$pp. Overall coverage is essentially unchanged, but the calibration is more honest: easy tasks no longer carry the calibration burden of the hard tasks.

\begin{table}[t]
\centering
\caption{Mondrian CP vs.\ standard R2CCP by task-difficulty group (LLaVA-Critic-7B, boundary-adjusted, 10 seeds).}
\label{tab:mondrian_app}
\small
\begin{tabular}{lcccc}
\toprule
& \multicolumn{2}{c}{\textbf{Standard R2CCP}} & \multicolumn{2}{c}{\textbf{Mondrian R2CCP}} \\
\cmidrule(lr){2-3} \cmidrule(lr){4-5}
\textbf{Group} & \textbf{Cov.} & \textbf{Width} & \textbf{Cov.} & \textbf{Width} \\
\midrule
Easy & 99.0\% & 3.55 & 97.5\% & \textbf{2.96} ($-16.6\%$) \\
Medium & 99.1\% & 3.60 & 98.7\% & \textbf{3.49} ($-3.1\%$) \\
Hard & 96.7\% & 3.63 & \textbf{98.2\%} & 3.78 ($+4.1\%$) \\
\midrule
Overall & 98.1\% & 3.60 & 98.2\% & \textbf{3.47} ($-3.6\%$) \\
\bottomrule
\end{tabular}
\end{table}

\subsection{Naive Split CP vs.\ R2CCP per dataset}
\label{app:naive_vs_r2ccp}

Table~\ref{tab:naive_vs_r2ccp} compares Naive Split CP and R2CCP per dataset. R2CCP produces strictly narrower intervals on all $14$ datasets, with the largest gain on MM-Vet ($-1.21$). The gain is largest on datasets where the residual distribution is most non-Gaussian (heavy tails or multi-modal), since Naive Split CP only models a single residual scale while R2CCP captures the full distribution.

\begin{table}[t]
\centering
\caption{Per-dataset comparison of Naive Split CP and R2CCP (raw width, LLaVA-Critic-7B, 10 seeds). R2CCP improves on every dataset.}
\label{tab:naive_vs_r2ccp}
\small
\begin{tabular}{lccc}
\toprule
\textbf{Dataset} & \textbf{Naive} & \textbf{R2CCP} & $\Delta$ \\
\midrule
AesBench & 2.53 & \textbf{2.08} & $-0.45$ \\
MM-Vet & 3.39 & \textbf{2.18} & $-1.21$ \\
WIT & 2.92 & \textbf{2.38} & $-0.54$ \\
COCO & 2.86 & \textbf{2.43} & $-0.43$ \\
Mind2Web & 3.13 & \textbf{2.69} & $-0.44$ \\
Conc.\ Cap. & 3.11 & \textbf{2.70} & $-0.41$ \\
TextVQA & 3.19 & \textbf{2.81} & $-0.38$ \\
LLaVA-B. & 3.28 & \textbf{2.92} & $-0.36$ \\
VisitBench & 3.37 & \textbf{2.96} & $-0.41$ \\
ChartQA & 3.49 & \textbf{3.08} & $-0.41$ \\
ScienceQA & 3.71 & \textbf{3.27} & $-0.44$ \\
MathVista & 3.64 & \textbf{3.37} & $-0.27$ \\
DiffusionDB & 3.56 & \textbf{3.41} & $-0.15$ \\
Infograph. & 3.67 & \textbf{3.50} & $-0.17$ \\
\midrule
\textbf{Average} & 3.27 & \textbf{2.84} & $-0.42$ \\
\bottomrule
\end{tabular}
\end{table}

\subsection{Effect of chain-of-thought prompting}
\label{app:cot}

Table~\ref{tab:cot_effect} compares the generic prompt (no CoT analysis step) and our CoT prompt for LLaVA-Critic-7B. CoT improves every correlation metric by $5$--$13\%$ and, more critically for conformal prediction, halves the rate of overconfident predictions (softmax max $> 0.99$, from $51.4\%$ to $25.6\%$) and reduces the extreme-overconfidence rate (max $> 0.999$) by $80\%$. The latter is the mechanism by which CoT enables tighter intervals: a less peaked softmax leaves discriminative signal in the score-token logprobs that R2CCP can exploit.

\begin{table}[t]
\centering
\caption{Effect of chain-of-thought prompting on LLaVA-Critic-7B.}
\label{tab:cot_effect}
\small
\begin{tabular}{lcc}
\toprule
\textbf{Metric} & \textbf{Generic prompt} & \textbf{CoT prompt} \\
\midrule
Pearson $\rho$ & .380 & \textbf{.402} ($+5.8\%$) \\
Spearman $\rho_s$ & .314 & \textbf{.356} ($+13.4\%$) \\
Kendall $\tau$ & .270 & \textbf{.300} ($+11.1\%$) \\
Exact accuracy & \textbf{33.9\%} & 32.2\% \\
MAE & 1.058 & \textbf{1.031} \\
\midrule
Overconf.\ ($> 0.99$) & 51.4\% & \textbf{25.6\%} ($-50\%$) \\
Overconf.\ ($> 0.999$) & 33.2\% & \textbf{6.6\%} ($-80\%$) \\
\bottomrule
\end{tabular}
\end{table}

\subsection{Seed sensitivity}
\label{app:seeds}

Table~\ref{tab:ablation_seeds} reports R2CCP coverage and width as a function of the number of calibration/test split seeds. Mean coverage and width are virtually constant from $5$ to $30$ seeds; only the standard deviation tightens modestly. Our reporting of $10$ seeds throughout the paper is therefore a conservative balance of compute and statistical stability.

\begin{table}[t]
\centering
\caption{R2CCP stability as a function of number of seeds (LLaVA-Critic-7B, MLLM-Judge). Means stabilize by 5 seeds, standard deviations tighten through 30.}
\label{tab:ablation_seeds}
\small
\begin{tabular}{rcccc}
\toprule
\textbf{Seeds} & \textbf{Coverage} & \textbf{$\pm$ std} & \textbf{Width} & \textbf{$\pm$ std} \\
\midrule
5 & .901 & .016 & 3.053 & .087 \\
10 & .900 & .016 & 3.049 & .097 \\
15 & .903 & .014 & 3.064 & .084 \\
20 & .902 & .014 & 3.058 & .085 \\
25 & .902 & .014 & 3.056 & .079 \\
30 & .902 & .013 & 3.058 & .074 \\
\bottomrule
\end{tabular}
\end{table}

\subsection{Calibration set size}
\label{app:split}

Table~\ref{tab:ablation_split} reports R2CCP performance as the calibration/test split varies from 70/30 to 30/70. Coverage stays within $1$pp of the $90\%$ target and width is essentially flat down to a 30/70 split (calibration set size $\sim 1{,}700$ samples). This robustness is consistent with the standard finite-sample CP guarantee, which requires only that the calibration quantile estimate not be too noisy. The result implies that practitioners can deploy CP for VLM judges with calibration sets considerably smaller than $5{,}000$ without a meaningful loss of validity.

\begin{table}[t]
\centering
\caption{R2CCP robustness to calibration/test split ratio (LLaVA-Critic-7B, MLLM-Judge, 10 seeds).}
\label{tab:ablation_split}
\small
\begin{tabular}{lrrcc}
\toprule
\textbf{Split} & \textbf{Cal N} & \textbf{Test N} & \textbf{Coverage} & \textbf{Width} \\
\midrule
70/30 & 4{,}001 & 1{,}715 & .900$\pm$.008 & 3.05$\pm$.04 \\
60/40 & 3{,}430 & 2{,}286 & .898$\pm$.017 & 3.05$\pm$.09 \\
50/50 & 2{,}858 & 2{,}858 & .900$\pm$.016 & 3.05$\pm$.10 \\
40/60 & 2{,}286 & 3{,}430 & .896$\pm$.020 & 3.06$\pm$.15 \\
30/70 & 1{,}715 & 4{,}001 & .887$\pm$.015 & 2.98$\pm$.10 \\
\bottomrule
\end{tabular}
\end{table}

\section{Polaris detailed analysis}
\label{app:polaris_full}

\subsection{Polaris confusion matrix}

Table~\ref{tab:polaris_confusion} reports the confusion matrix for LLaVA-Critic-7B on Polaris. The judge achieves near-perfect identification of GT $= 1$ captions ($99.6\%$ exact accuracy), demonstrating that on a well-defined captioning task the judge can decisively flag clearly bad outputs. Most remaining confusion is between adjacent Likert levels.

\begin{table}[t]
\centering
\caption{Confusion matrix for LLaVA-Critic-7B on Polaris (\%, GT rows, Pred columns). GT $= 1$ is near-perfect (99.6\%).}
\label{tab:polaris_confusion}
\small
\begin{tabular}{l|ccccc|c}
\toprule
& \textbf{P=1} & \textbf{P=2} & \textbf{P=3} & \textbf{P=4} & \textbf{P=5} & $\pm$\textbf{1} \\
\midrule
GT=1 & 99.6 & 0.1 & 0.2 & 0.1 & 0.0 & 99.7 \\
GT=2 & 74.2 & 12.9 & 5.7 & 4.8 & 2.4 & 92.7 \\
GT=3 & 18.6 & 13.4 & 32.1 & 30.8 & 5.2 & 76.2 \\
GT=4 & 1.6 & 4.0 & 14.6 & 64.2 & 15.6 & 94.4 \\
GT=5 & 0.8 & 2.0 & 5.6 & 63.9 & 27.8 & 91.7 \\
\bottomrule
\end{tabular}
\end{table}

\subsection{Polaris error-bin CP coverage}

Table~\ref{tab:polaris_error_bins} reports CP coverage by error magnitude on Polaris. The intervals here are an order of magnitude tighter (mean width $0.68$) than on MLLM-Judge, so $\pm 1$ judge errors no longer fit comfortably inside the raw interval (raw coverage $58.1\%$). Boundary adjustment expands the interval to integer endpoints and recovers $\pm 1$ coverage to $99.4\%$. The same mechanism applies on MLLM-Judge but with much weaker effect because intervals there are already wide.

\begin{table}[t]
\centering
\caption{CP coverage by error magnitude on Polaris (R2CCP, LLaVA-Critic-7B). Tight intervals reduce raw coverage on $\pm 1$ errors; boundary adjustment recovers it to $99.4\%$.}
\label{tab:polaris_error_bins}
\small
\begin{tabular}{lrcc}
\toprule
\textbf{Error} & \textbf{\% samples} & \textbf{Cov. (raw)} & \textbf{Cov. (adj.)} \\
\midrule
Exact (0) & 80.9\% & 98.2\% & 99.6\% \\
$\pm 1$ & 14.6\% & 58.1\% & 99.4\% \\
$\pm 2$ & 4.1\% & 50.0\% & 80.7\% \\
$\pm 3$ & 0.5\% & 0.6\% & 52.4\% \\
\bottomrule
\end{tabular}
\end{table}

\subsection{Phi-4 on Polaris: model specialization vs.\ scale}
\label{app:phi4_polaris}

Table~\ref{tab:phi4_polaris} compares LLaVA-Critic-7B and Phi-4-15B on the Polaris captioning benchmark. Despite Phi-4 being twice as large in parameter count, LLaVA-Critic dominates on every metric: $7.1\times$ higher Pearson, $3.1\times$ higher exact accuracy, $5.3\times$ narrower R2CCP interval. With reasoning-mode disabled (\texttt{enable\_thinking=False}), Phi-4 collapses to score 4 for $72\%$ of samples and Pearson drops to $0.036$. The interpretation is direct: \emph{evaluation specialization matters more than scale}. LLaVA-Critic was trained explicitly to act as a multimodal judge; Phi-4 was not. On a domain that exercises that specialization (single, well-defined captioning), the gap is enormous. We expect this gap to narrow on more general tasks where neither model has explicit fine-tuning advantage.

\begin{table}[t]
\centering
\caption{Judge comparison on Polaris. LLaVA-Critic (7B, evaluation-specialized) outperforms Phi-4 (15B, general-purpose) by a wide margin on every metric.}
\label{tab:phi4_polaris}
\small
\begin{tabular}{lccc}
\toprule
& \textbf{LLaVA-Critic} & \textbf{Phi-4} & \textbf{Phi-4} \\
& \textbf{7B} & \textbf{15B (think)} & \textbf{15B (no-think)} \\
\midrule
Samples & 8{,}726 & 924 & 8{,}726 \\
Pearson & \textbf{.906} & .128 & .036 \\
Accuracy & \textbf{80.9\%} & 26.1\% & 20.0\% \\
$\pm 1$ Acc. & \textbf{95.4\%} & 60.5\% & 37.5\% \\
R2CCP Width & \textbf{0.68} & 3.60 & 2.96 \\
\bottomrule
\end{tabular}
\end{table}

\section{Multi-judge feature fusion}
\label{app:multijudge_full}

We test whether stacking score-token logprob features from multiple judges into a single feature vector improves R2CCP performance. The fused feature vectors have dimension $5k$ for $k$ judges combined. Table~\ref{tab:multijudge_app} (reproducing the main-text Table~\ref{tab:multijudge}) shows that feature fusion strictly underperforms the best single judge: the all-three-judge fusion (15-dim features) yields width $3.14$, wider than Gemini alone ($2.85$). The same pattern holds for two-judge combinations.

The mechanism is straightforward. R2CCP's nonconformity mapping is learned from the calibration set, which contains roughly $2{,}900$ samples in the 50/50 split. Increasing the input dimension from $5$ to $10$ or $15$ without proportionally increasing calibration data adds noise from the weaker judges' logprobs without sufficient signal to overcome it. The result is a learned nonconformity that overfits the calibration set and generalizes worse on the test set. The takeaway is that for VLM-judge conformal prediction, \emph{feature quality dominates feature quantity}: a single strong judge with well-calibrated logprobs is preferable to ensembling logprobs from multiple judges of mixed quality.

\begin{table}[t]
\centering
\caption{Multi-judge feature fusion (R2CCP, 10 seeds). Combining judges does not outperform the best single judge (Gemini). Logprob quality $>$ quantity.}
\label{tab:multijudge_app}
\small
\begin{tabular}{lrccc}
\toprule
\textbf{Configuration} & \textbf{Feat.} & \textbf{Cov. (raw)} & \textbf{Width (raw)} & \textbf{Width (adj.)} \\
\midrule
Gemini only & 5 & .898 & \textbf{2.85} & \textbf{3.41} \\
LLaVA + Gemini & 10 & .901 & 3.02 & 3.51 \\
LLaVA only & 5 & .900 & 3.05 & 3.60 \\
Multi-Judge (all 3) & 15 & .905 & 3.14 & 3.56 \\
LLaVA + Phi-4 & 10 & .903 & 3.18 & 3.62 \\
Phi-4 only & 5 & .891 & 3.13 & 3.70 \\
\bottomrule
\end{tabular}
\end{table}


\begin{thebibliography}{39}
\providecommand{\natexlab}[1]{#1}
\providecommand{\url}[1]{\texttt{#1}}
\expandafter\ifx\csname urlstyle\endcsname\relax
  \providecommand{\doi}[1]{doi: #1}\else
  \providecommand{\doi}{doi: \begingroup \urlstyle{rm}\Url}\fi

\bibitem[Abdin et~al.(2025)Abdin, Agarwal, Agrawal, et~al.]{abdin2025phi4reasoning}
Marah Abdin, Sahil Agarwal, Aman Agrawal, et~al.
\newblock Phi-4-reasoning technical report.
\newblock \emph{arXiv preprint arXiv:2504.21318}, 2025.

\bibitem[Angelopoulos et~al.(2021)Angelopoulos, Bates, Cand{\`e}s, Jordan, and Lei]{angelopoulos2021conformal}
Anastasios~N Angelopoulos, Stephen Bates, Emmanuel~J Cand{\`e}s, Michael~I Jordan, and Lihua Lei.
\newblock Learn then test: Calibrating predictive algorithms to achieve risk control.
\newblock \emph{arXiv preprint arXiv:2110.01052}, 2021.

\bibitem[Darabi et~al.(2024)Darabi, Shukla, Jayasuriya, Kumar, Stutts, and Trivedi]{darabi2024navigating}
Nastaran Darabi, Priyesh Shukla, Dinithi Jayasuriya, Divake Kumar, Alex~Christopher Stutts, and Amit~Ranjan Trivedi.
\newblock Navigating the unknown: Uncertainty-aware compute-in-memory autonomy of edge robotics.
\newblock In \emph{2024 Design, Automation \& Test in Europe Conference \& Exhibition (DATE)}, pp.\  1--6, 2024.

\bibitem[Chen et~al.(2024)Chen, Chen, Zhang, Liu, Wang, Zhou, Zhang, Wan, Zhou, and Sun]{chen2024mllmasajudge}
Dongping Chen, Ruoxi Chen, Shilin Zhang, Yinuo Liu, Yaochen Wang, Huichi Zhou, Qihui Zhang, Yao Wan, Pan Zhou, and Lichao Sun.
\newblock {MLLM}-as-a-judge: Assessing multimodal {LLM}-as-a-judge with vision-language benchmark.
\newblock In \emph{International Conference on Machine Learning (ICML)}, 2024.

\bibitem[Kumar et~al.(2025{\natexlab{a}})Kumar, Poggi, Tayebati, Naik, Ahuja, and Trivedi]{kumar2025calibrated}
Divake Kumar, Patrick Poggi, Sina Tayebati, Devashri Naik, Nilesh Ahuja, and Amit~Ranjan Trivedi.
\newblock Calibrated decomposition of aleatoric and epistemic uncertainty in deep features for inference-time adaptation.
\newblock \emph{arXiv preprint arXiv:2511.12389}, 2025{\natexlab{a}}.

\bibitem[{Google DeepMind}(2025)]{google2025gemini25}
{Google DeepMind}.
\newblock Gemini 2.5: Our newest {Gemini} model with thinking.
\newblock \url{https://blog.google/technology/google-deepmind/gemini-model-thinking-updates-march-2025/}, 2025.

\bibitem[Tayebati et~al.(2025{\natexlab{a}})Tayebati, Kumar, Darabi, Jayasuriya, Krishnan, and Trivedi]{tayebati2025conformalabstention}
Sina Tayebati, Divake Kumar, Nastaran Darabi, Dinithi Jayasuriya, Ranganath Krishnan, and Amit~Ranjan Trivedi.
\newblock Learning conformal abstention policies for adaptive risk management in large language and vision-language models.
\newblock \emph{arXiv preprint arXiv:2502.06884}, 2025{\natexlab{a}}.

\bibitem[Guha et~al.(2024)Guha, Natarajan, M{\"o}llenhoff, Khan, and Ndiaye]{guha2024conformal}
Etash Guha, Shlok Natarajan, Thomas M{\"o}llenhoff, Mohammad~Emtiyaz Khan, and Eugene Ndiaye.
\newblock Conformal prediction via regression-as-classification.
\newblock In \emph{International Conference on Learning Representations (ICLR)}, 2024.

\bibitem[He et~al.(2024)He, Zhang, and Roth]{he2024soceval}
Hangfeng He, Hongming Zhang, and Dan Roth.
\newblock {SocREval}: Large language models with the socratic method for reference-free reasoning evaluation.
\newblock In \emph{Findings of the Association for Computational Linguistics: NAACL 2024}, 2024.

\bibitem[Stutts et~al.(2024)Stutts, Kumar, Tulabandhula, and Trivedi]{stutts2024conformal}
Alex~Christopher Stutts, Divake Kumar, Theja Tulabandhula, and Amit~Ranjan Trivedi.
\newblock Conformal inference meets evidential learning: Distribution-free uncertainty quantification with epistemic and aleatoric separability.
\newblock In \emph{Proceedings of the 61st ACM/IEEE Design Automation Conference (DAC)}, pp.\  1--4, 2024.

\bibitem[Kumar et~al.(2023)Kumar, Lu, Gupta, Palepu, Bellamy, Raskar, and Beam]{kumar2023conformal}
Bhawesh Kumar, Charlie Lu, Gauri Gupta, Anil Palepu, David Bellamy, Ramesh Raskar, and Andrew Beam.
\newblock Conformal prediction with large language models for multi-choice question answering.
\newblock \emph{arXiv preprint arXiv:2305.18404}, 2023.

\bibitem[Li et~al.(2025)Li, Wei, Xie, Yang, Song, Wang, An, Liu, Li, Lin, Kong, and Liu]{li2024vlrewardbench}
Lei Li, Yuancheng Wei, Zhihui Xie, Xuqing Yang, Yifan Song, Peiyi Wang, Chenxin An, Tianyu Liu, Sujian Li, Bill~Yuchen Lin, Lingpeng Kong, and Qi~Liu.
\newblock {VL-RewardBench}: A challenging benchmark for vision-language generative reward models.
\newblock In \emph{Proceedings of the IEEE/CVF Conference on Computer Vision and Pattern Recognition (CVPR)}, 2025.

\bibitem[Darabi et~al.(2025)Darabi, Kumar, Tayebati, and Trivedi]{darabi2025intact}
Nastaran Darabi, Divake Kumar, Sina Tayebati, and Amit~Ranjan Trivedi.
\newblock {INTACT}: Inducing noise tolerance through adversarial curriculum training for {LiDAR}-based safety-critical perception and autonomy.
\newblock \emph{arXiv preprint arXiv:2502.01896}, 2025.

\bibitem[Lin et~al.(2021)Lin, Trivedi, and Sun]{lin2021locally}
Zhen Lin, Shubhendu Trivedi, and Jimeng Sun.
\newblock Locally valid and discriminative prediction intervals for deep learning models.
\newblock In \emph{Advances in Neural Information Processing Systems}, 2021.

\bibitem[Kumar et~al.(2025{\natexlab{b}})Kumar, Tayebati, Darabi, Hu, and Trivedi]{kumar2025lidarcp}
Divake Kumar, Sina Tayebati, Nastaran Darabi, Vita Pi-Ho Hu, and Amit~Ranjan Trivedi.
\newblock Uncertainty-aware {LiDAR}-camera autonomy via conformal prediction and principled abstention.
\newblock In \emph{2025 IEEE International Conference on Omni-layer Intelligent Systems (COINS)}, pp.\  1--6. IEEE, 2025{\natexlab{b}}.
\newblock \doi{10.1109/COINS65080.2025.11125785}.

\bibitem[Liu et~al.(2023)Liu, Iter, Xu, Wang, Xu, and Zhu]{liu2023geval}
Yang Liu, Dan Iter, Yichong Xu, Shuohang Wang, Ruochen Xu, and Chenguang Zhu.
\newblock {G-Eval}: {NLG} evaluation using {GPT-4} with better human alignment.
\newblock In \emph{Proceedings of the 2023 Conference on Empirical Methods in Natural Language Processing (EMNLP)}, pp.\  2511--2522, 2023.

\bibitem[Lu et~al.(2022)Lu, Lemay, Chang, H{\"o}bel, and Kalpathy-Cramer]{lu2022fair}
Charles Lu, Andr{\'e}anne Lemay, Ken Chang, Katharina H{\"o}bel, and Jayashree Kalpathy-Cramer.
\newblock Fair conformal predictors for applications in medical imaging.
\newblock In \emph{Proceedings of the AAAI Conference on Artificial Intelligence}, volume~36, pp.\  12008--12016, 2022.

\bibitem[Naik et~al.(2026)Naik, Kumar, Darabi, and Trivedi]{naik2026belief}
Devashri Naik, Divake Kumar, Nastaran Darabi, and Amit~Ranjan Trivedi.
\newblock Belief dynamics for detecting behavioral shifts in safe collaborative manipulation.
\newblock \emph{arXiv preprint arXiv:2604.04967}, 2026.

\bibitem[Mohri \& Hashimoto(2024)Mohri and Hashimoto]{mohri2024language}
Christopher Mohri and Tatsunori Hashimoto.
\newblock Language models with conformal factuality guarantees.
\newblock In \emph{International Conference on Machine Learning (ICML)}, 2024.

\bibitem[Poggi et~al.(2026)Poggi, Kumar, Tulabandhula, and Trivedi]{poggi2026depth}
Patrick Poggi, Divake Kumar, Theja Tulabandhula, and Amit~Ranjan Trivedi.
\newblock Uncertainty-guided inference-time depth adaptation for transformer-based visual tracking.
\newblock \emph{arXiv preprint arXiv:2602.16160}, 2026.

\bibitem[Quach et~al.(2024)Quach, Fisch, Schuster, Yala, Sohn, Jaakkola, and Barzilay]{quach2024conformal}
Victor Quach, Adam Fisch, Tal Schuster, Adam Yala, Jae~Ho Sohn, Tommi~S Jaakkola, and Regina Barzilay.
\newblock Conformal language modeling.
\newblock In \emph{International Conference on Learning Representations (ICLR)}, 2024.

\bibitem[Tayebati et~al.(2025{\natexlab{b}})Tayebati, Kumar, Darabi, Jayasuriya, Tulabandhula, Krishnan, and Trivedi]{tayebati2025cap}
Sina Tayebati, Divake Kumar, Nastaran Darabi, Dinithi Jayasuriya, Theja Tulabandhula, Ranganath Krishnan, and Amit~Ranjan Trivedi.
\newblock {CAP}: Conformalized abstention policies for context-adaptive risk management for {LLMs} and {VLMs}.
\newblock In \emph{Proceedings of the 17th Asian Conference on Machine Learning (ACML), Conference Track}, 2025{\natexlab{b}}.

\bibitem[Romano et~al.(2019)Romano, Patterson, and Cand{\`e}s]{romano2019conformalized}
Yaniv Romano, Evan Patterson, and Emmanuel Cand{\`e}s.
\newblock Conformalized quantile regression.
\newblock In \emph{Advances in Neural Information Processing Systems}, 2019.

\bibitem[Kumar et~al.(2025{\natexlab{c}})Kumar, Tayebati, Migliarba, Krishnan, and Trivedi]{kumar2025learnablecp}
Divake Kumar, Sina Tayebati, Francesco Migliarba, Ranganath Krishnan, and Amit~Ranjan Trivedi.
\newblock Learnable conformal prediction with context-aware nonconformity functions for robotic planning and perception.
\newblock \emph{arXiv preprint arXiv:2509.21955}, 2025{\natexlab{c}}.

\bibitem[Sesia \& Cand{\`e}s(2020)Sesia and Cand{\`e}s]{sesia2020comparison}
Matteo Sesia and Emmanuel~J Cand{\`e}s.
\newblock A comparison of some conformal quantile regression methods.
\newblock \emph{Stat}, 9\penalty0 (1):\penalty0 e261, 2020.

\bibitem[Sesia \& Romano(2021)Sesia and Romano]{sesia2021conformal}
Matteo Sesia and Yaniv Romano.
\newblock Conformal prediction using conditional histograms.
\newblock In \emph{Advances in Neural Information Processing Systems}, 2021.

\bibitem[Trivedi et~al.(2025)Trivedi, Tayebati, Kumawat, Darabi, Kumar, Kosta, Venkatesha, Jayasuriya, Jayasinghe, Panda, Mukhopadhyay, and Roy]{trivedi2025sensing}
Amit~Ranjan Trivedi, Sina Tayebati, Hemant Kumawat, Nastaran Darabi, Divake Kumar, Adarsh~Kumar Kosta, Yeshwanth Venkatesha, Dinithi Jayasuriya, Nethmi Jayasinghe, Priyadarshini Panda, Saibal Mukhopadhyay, and Kaushik Roy.
\newblock Intelligent sensing-to-action for robust autonomy at the edge: Opportunities and challenges.
\newblock In \emph{2025 Design, Automation \& Test in Europe Conference (DATE)}, pp.\  1--10, 2025.

\bibitem[Sheng et~al.(2025)Sheng, Liu, He, Zhao, and Kang]{sheng2025analyzing}
Huanxin Sheng, Xinyi Liu, Hangfeng He, Jieyu Zhao, and Jian Kang.
\newblock Analyzing uncertainty of {LLM}-as-a-judge: Interval evaluations with conformal prediction.
\newblock In \emph{Proceedings of the 2025 Conference on Empirical Methods in Natural Language Processing (EMNLP)}, pp.\  11286--11328, 2025.

\bibitem[Su et~al.(2024)Su, Luo, Wang, and Cheng]{su2024api}
Jiayuan Su, Jing Luo, Hongwei Wang, and Lu~Cheng.
\newblock {API} is enough: Conformal prediction for large language models without logit-access.
\newblock In \emph{Findings of the Association for Computational Linguistics: EMNLP 2024}, 2024.

\bibitem[Tayebati et~al.(2026)Tayebati, Kumar, Darabi, Ettori, Krishnan, and Trivedi]{tayebati2026tracer}
Sina Tayebati, Divake Kumar, Nastaran Darabi, Davide Ettori, Ranganath Krishnan, and Amit~Ranjan Trivedi.
\newblock {TRACER}: Trajectory risk aggregation for critical episodes in agentic reasoning.
\newblock \emph{arXiv preprint arXiv:2602.11409}, 2026.

\bibitem[Tian et~al.(2023)Tian, Mitchell, Zhou, Sharma, Rafailov, Yao, Finn, and Manning]{tian2023consistency}
Katherine Tian, Eric Mitchell, Allan Zhou, Archit Sharma, Rafael Rafailov, Huaxiu Yao, Chelsea Finn, and Christopher~D Manning.
\newblock Just ask for calibration: Strategies for eliciting calibrated confidence scores from language models fine-tuned with human feedback.
\newblock In \emph{Proceedings of the 2023 Conference on Empirical Methods in Natural Language Processing (EMNLP)}, 2023.

\bibitem[Vovk et~al.(2005)Vovk, Gammerman, and Shafer]{vovk2005algorithmic}
Vladimir Vovk, Alexander Gammerman, and Glenn Shafer.
\newblock \emph{Algorithmic Learning in a Random World}.
\newblock Springer, 2005.

\bibitem[Kumar et~al.(2026)Kumar, Tayebati, Naik, Poggi, Rios, Ahuja, and Trivedi]{kumar2026triage}
Divake Kumar, Sina Tayebati, Devashri Naik, Patrick Poggi, Amanda~Sofie Rios, Nilesh Ahuja, and Amit~Ranjan Trivedi.
\newblock {TRIAGE}: Type-routed interventions via aleatoric-epistemic gated estimation in robotic manipulation and adaptive perception---don't treat all uncertainty the same.
\newblock \emph{arXiv preprint arXiv:2603.08128}, 2026.

\bibitem[Wada et~al.(2024)Wada, Kaneda, Saito, and Sugiura]{wada2024polos}
Yuiga Wada, Kanta Kaneda, Daichi Saito, and Komei Sugiura.
\newblock Polos: Multimodal metric learning from human feedback for image captioning.
\newblock In \emph{Proceedings of the IEEE/CVF Conference on Computer Vision and Pattern Recognition (CVPR)}, 2024.

\bibitem[Wagner et~al.(2024)Wagner, Desmond, Nair, Ashktorab, Daly, Pan, Cooper, Johnson, and Geyer]{wagner2024llmconfidence}
Nico Wagner, Michael Desmond, Rahul Nair, Zahra Ashktorab, Elizabeth~M Daly, Qian Pan, Martin~Santillan Cooper, James~M Johnson, and Werner Geyer.
\newblock Black-box uncertainty quantification method for {LLM}-as-a-judge.
\newblock \emph{arXiv preprint arXiv:2410.11594}, 2024.

\bibitem[Xie et~al.(2024)Xie, Barber, and Cand{\`e}s]{xie2024boosted}
Ran Xie, Rina~Foygel Barber, and Emmanuel~J Cand{\`e}s.
\newblock Boosted conformal prediction intervals.
\newblock In \emph{Advances in Neural Information Processing Systems}, 2024.

\bibitem[Xiong et~al.(2024)Xiong, Hu, Lu, Li, Fu, He, and Hooi]{xiong2024selfconsistency}
Miao Xiong, Zhiyuan Hu, Xinyang Lu, Yifei Li, Jie Fu, Junxian He, and Bryan Hooi.
\newblock Can {LLMs} express their uncertainty? an empirical evaluation of confidence elicitation in {LLMs}.
\newblock In \emph{International Conference on Learning Representations (ICLR)}, 2024.

\bibitem[Xiong et~al.(2025)Xiong, Wang, Guo, Ye, Fan, Gu, Huang, and Li]{xiong2024llavacritic}
Tianyi Xiong, Xiyao Wang, Dong Guo, Qinghao Ye, Haoqi Fan, Quanquan Gu, Heng Huang, and Chunyuan Li.
\newblock {LLaVA-Critic}: Learning to evaluate multimodal models.
\newblock In \emph{Proceedings of the IEEE/CVF Conference on Computer Vision and Pattern Recognition (CVPR)}, 2025.

\bibitem[Zheng et~al.(2023)Zheng, Chiang, Sheng, Zhuang, Wu, Zhuang, Lin, Li, Li, Xing, et~al.]{zheng2023judging}
Lianmin Zheng, Wei-Lin Chiang, Ying Sheng, Siyuan Zhuang, Zhanghao Wu, Yonghao Zhuang, Zi~Lin, Zhuohan Li, Dacheng Li, Eric~P Xing, et~al.
\newblock Judging {LLM}-as-a-judge with {MT-Bench} and chatbot arena.
\newblock In \emph{Advances in Neural Information Processing Systems}, 2023.

\end{thebibliography}
\end{document}